\documentclass[sn-mathphys,Numbered]{sn-jnl}

\usepackage{graphicx}%
\usepackage{multirow}%
\usepackage{amsmath,amssymb,amsfonts}%
\usepackage{amsthm}%
\usepackage{mathrsfs}%
\usepackage[title]{appendix}%
\usepackage[dvipsnames]{xcolor}%
\usepackage{textcomp}%
\usepackage{manyfoot}%
\usepackage{booktabs}%
\usepackage{algorithm}%
\usepackage{algorithmicx}%
\usepackage{algpseudocode}%
\usepackage{listings}%
\usepackage{wrapfig}
\usepackage{blkarray, bigstrut} 

\theoremstyle{thmstyleone}%
\theoremstyle{thmstyletwo}%

\theoremstyle{thmstylethree}%
\usepackage{blkarray}
\usepackage{makecell}
\usepackage{bbding}
\usepackage{listings}
\usepackage{xcolor}
\PassOptionsToPackage{usenames, dvipsnames}{xcolor}
\definecolor{codegreen}{rgb}{0,0.6,0}
\definecolor{codegray}{rgb}{0.5,0.5,0.5}
\definecolor{codepurple}{rgb}{0.58,0,0.82}
\definecolor{backcolour}{rgb}{0.95,0.95,0.92}
\definecolor{indianred}{rgb}{0.9,0.3608,0.3608}

\lstdefinestyle{mystyle}{
    backgroundcolor=\color{backcolour},   
    commentstyle=\color{codegreen},
    keywordstyle=\color{black},
    numberstyle=\tiny\color{codegray},
    stringstyle=\color{codepurple},
    basicstyle=\footnotesize,
    breakatwhitespace=false,         
    breaklines=true,                 
    captionpos=b,                    
    keepspaces=true,                 
    numbers=left,                    
    numbersep=5pt,                  
    showspaces=false,                
    showstringspaces=false,
    showtabs=false,                  
    tabsize=2
}
\usepackage{listings}
\usepackage{xcolor}
\definecolor{codegreen}{rgb}{0,0.6,0}
\definecolor{codegray}{rgb}{0.5,0.5,0.5}
\definecolor{codepurple}{rgb}{0.58,0,0.82}
\definecolor{backcolour}{rgb}{0.95,0.95,0.92}
\lstdefinestyle{mystyle}{
    backgroundcolor=\color{backcolour},   
    commentstyle=\color{codegreen},
    keywordstyle=\color{black},
    numberstyle=\tiny\color{codegray},
    stringstyle=\color{codepurple},
    basicstyle=\footnotesize,
    breakatwhitespace=false,         
    breaklines=true,                 
    captionpos=b,                    
    keepspaces=true,                 
    numbers=left,                    
    numbersep=5pt,                  
    showspaces=false,                
    showstringspaces=false,
    showtabs=false,                  
    tabsize=2
}

\newcommand{\eg}{\emph{e.g.,}~} 

\newcommand{\ie}{\emph{i.e.,}~}

\begin{document}

\title[Article Title]{Deep Meta Programming}
\title[Article Title]{Neural Meta-Symbolic Reasoning and Learning}

\author*[1,3]{\fnm{Zihan} \sur{Ye}}\email{zihan.ye@tu-darmstadt.de}

\author[1]{\fnm{Hikaru} \sur{Shindo}}\email{hikaru.shindo@tu-darmstadt.de}

\author[3,5]{\fnm{Devendra Singh} \sur{Dhami}}\email{d.s.dhami@tue.nl}

\author[1,2,3,4]{\fnm{Kristian} \sur{Kersting}}\email{kersting@cs.tu-darmstadt.de}

\affil[1]{\orgdiv{AI and Machine Learning Group, Dept. of Computer Science}, \orgname{TU Darmstadt},  \country{Germany}}

\affil[2]{\orgdiv{Centre for Cognitive Science}, \orgname{TU Darmstadt},\country{Germany}}

\affil[3]{\orgdiv{Hessian Center for AI (hessian.AI)}, \country{Germany}}
\affil[4]{\orgdiv{German Center for Artificial Intelligence (DFKI)}, \country{Germany}}
\affil[5]{\orgdiv{Eindhoven University of Technology}, \country{Netherlands}}


\abstract{Deep neural learning uses an increasing amount of computation and data to solve very specific problems. By stark contrast, 
human minds solve a wide range of problems using a fixed amount of computation and limited experience. One
ability that seems crucial to this kind of general intelligence is meta-reasoning, \ie our ability to reason about reasoning. To make deep learning do more from less, we propose the first neural \textbf{meta}-symbolic system (NEMESYS) for reasoning and learning: meta programming using differentiable forward-chaining reasoning in first-order logic. Differentiable meta programming naturally allows NEMESYS to reason and learn several tasks efficiently. 
This is different from performing object-level deep reasoning and learning, which refers in some way to entities external to the system. In contrast, NEMESYS enables self-introspection, lifting from object- to meta-level reasoning and vice versa. 
In our extensive experiments, we demonstrate that NEMESYS can solve different kinds of tasks by adapting the meta-level programs without modifying the internal reasoning system. Moreover, we show that NEMESYS can learn meta-level programs given examples.
This is difficult, if not impossible, for standard differentiable logic programming. }




\maketitle

\begin{figure}[h]
    \centering
    \includegraphics[trim=0 80 00 82, clip, width=\textwidth]{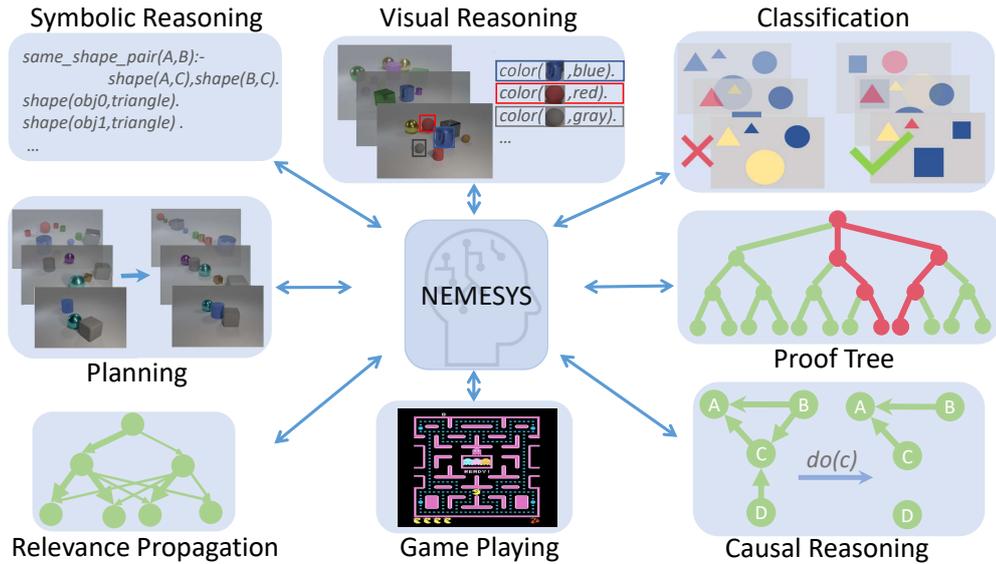}
    \caption{\textbf{NEMESYS solves different kinds of tasks by using meta-level reasoning and learning.}  NEMESYS addresses, for instance, visual reasoning, planning, and causal reasoning without modifying its internal reasoning architecture.
    (Best viewed in color)} 
    \label{fig:meta_reason_compare}
\end{figure}
\section{Introduction}\label{sec1}
One of the distant goals of Artificial Intelligence (AI) is to build a fully autonomous or ‘human-like’ system. The current successes of deep learning systems such as DALLE-2~\citep{dalle2}, ChatGPT~\citep{stiennon2020learning,floridi2020gpt}, and Gato~\citep{reedgeneralist} have been promoted as bringing the field closer to this goal. However, current systems still require a large number of computations and often solve rather specific tasks. For example, DALLE-2 can generate very high-quality images but cannot play chess or Atari games. In stark contrast, human minds solve a wide range of problems using a small amount of computation and limited experience. 

Most importantly, to be considered a major step towards achieving Artificial General Intelligence (AGI), a system must not only be able to perform a variety of tasks, such as Gato \citep{reedgeneralist} playing Atari games, captioning images, chatting, and controlling a real robot arm, but also be self-reflective and able to learn and reason about its own capabilities. This means that it must be able to improve itself and adapt to new situations through self-reflection \citep{ackerman2017metareasoning,Costantini2002meta,Griffiths2019meta,russell1991metaprinciples}.
Consequently, the study of meta-level architectures such as meta learning~\citep{schmidhuber1987evolutionary} and meta-reasoning~\citep{Griffiths2019meta} becomes progressively important.
Meta learning~\citep{Thrun1998learningtolearn} is a way to improve the learning algorithm itself~\citep{Finn17modelagnostic_metalearning,Hospedales22metalearning}, \ie it performs learning at a higher level, or \emph{meta-level}. 
Meta-reasoning is a related concept that involves a system being able to think about its own abilities and how it processes information~\citep{ackerman2017metareasoning,Costantini2002meta}. It involves reflecting on, or introspecting about, the system's own reasoning processes.


Indeed, meta-reasoning is different from object-centric reasoning, which refers to the system thinking about entities external to itself~\cite{kim2018notsoclevr,Stammer21clevrhans,Shindo21nsfr}. 
Here, the models perform low-level visual perception and reasoning on high-level concepts. Accordingly, there has been a push to make these reasoning systems differentiable~\citep{Evans18,Shindo21aaai} along with addressing benchmarks in a visual domain such as CLEVR~\citep{Johnson17clevr} and Kandinsky patterns~\citep{Holzinger19IQ,Mueller21}. They use object-centric neural networks to perceive objects and perform reasoning using their output.
Although this can solve the proposed benchmarks to some extent, the critical question remains unanswered:  \textit{Is the reasoner able to justify 
its own operations? Can the same model solve different tasks such as (causal) reasoning, planning, game playing, and much more? } 

To overcome these limitations, 
we propose NEMESYS,  the first neural \textbf{meta}-symbolic reasoning system. 
NEMESYS extensively performs \emph{meta-level} programming on neuro-symbolic systems, and thus it can reason and learn several tasks. 
This is different from performing object-level deep reasoning and learning, which refers in some way to entities external to the system. 
NEMESYS is able to reflect or introspect, \ie to shift from object- to meta-level reasoning and vice versa. 
\begin{table*}[t]
\small
\centering
\begin{tabular}{lcccc}
                  & Meta Reasoning  & Multitask Adaptation & \makecell{Differentiable Meta\\Structure Learning}\\ \hline
DeepProbLog~\citep{manhaeve2018deepproblog}        & \color{red}{\XSolidBrush}        &\color{red}{\XSolidBrush}           & \color{red}{\XSolidBrush}           \\
NTPs~\citep{rocktaschel2017end}         & \color{red}{\XSolidBrush}      & \color{red}{\XSolidBrush}           & \color{red}{\XSolidBrush}            \\
FFSNL~\citep{cunnington2023ffnsl}           &\color{red}{\XSolidBrush}        & \color{red}{\XSolidBrush}           &\color{red}{\XSolidBrush}             \\
$\alpha$ILP~\citep{shindo2023alphailp}              & \color{red}{\XSolidBrush}       &\color{red}{\XSolidBrush}              &\color{red}{\XSolidBrush}       \\
Scallop~\citep{huang2021scallop}            & \color{red}{\XSolidBrush}         & \color{red}{\XSolidBrush}           & \color{red}{\XSolidBrush}         \\
NeurASP~\citep{Yang2020neuroasp}           & \color{red}{\XSolidBrush}        & \color{red}{\XSolidBrush}           & \color{red}{\XSolidBrush}               \\
NEMESYS~(ours)  &\color{blue}{\Checkmark}      & \color{blue}{\Checkmark}         &\color{blue}{\Checkmark}           \\ \hline
\end{tabular}
\caption{\textbf{Comparisons between NEMESYS and other state-of-the-art Neuro-Symbolic systems.} We compare these systems with NEMESYS in three aspects, whether the system performs \textbf{meta reasoning}, can the same system \textbf{adapt to solve different tasks} and is the system capable of \textbf{differentiable meta level structure learning}.}
\label{tab:compare_tasks}
\end{table*}

Overall, we make the following contributions: 
\begin{enumerate}
    \item We propose NEMESYS, the first neural \textbf{meta}-symbolic reasoning and learning system that performs differentiable forward reasoning using meta-level programs.
    \item To evaluate the ability of NEMESYS, we propose a challenging task, \emph{visual concept repairing}, where the task is to rearrange objects in visual scenes based on relational logical concepts. 
    \item We empirically show that NEMESYS can efficiently solve different visual reasoning tasks with meta-level programs, achieving comparable performances with object-level forward reasoners~\citep{Evans18,shindo2023alphailp} that use specific programs for each task.
    \item Moreover, we empirically show that using powerful differentiable meta-level programming, NEMESYS can solve different kinds of tasks that are difficult, if not impossible, for the previous neuro-symbolic systems. In our experiments, NEMESYS provides the function of (i) \emph{reasoning with integrated proof generation}, \emph{i.e.,} performing differentiable reasoning producing proof trees,  (ii) \emph{explainable artificial intelligence (XAI)}, \emph{i.e.,} highlighting the importance of logical atoms given conclusions, (iii) \emph{reasoning avoiding infinite loops}, \emph{i.e.,} performing differentiable reasoning on programs which cause infinite loops that the previous logic reasoning systems unable to solve, and (iv) \emph{differentiable causal reasoning}, \emph{i.e.,} performing causal reasoning~\citep{pearl2009causality,pearl2012docalculus} on a causal Bayesian network using differentiable  meta reasoners. To the best of the authors' knowledge, we propose the first \emph{differentiable} $\mathtt{do}$ operator. Achieving these functions with object-level reasoners necessitates significant efforts, and in some cases, it may be unattainable. In stark contrast, NEMESYS successfully realized the different useful functions by having different meta-level programs \emph{without any modifications of the reasoning function itself}.
    \item We demonstrate that NEMESYS can perform structure learning on the meta-level, \ie learning meta programs from examples and adapting itself to solve different tasks automatically by learning efficiently with gradients. 
    
\end{enumerate}

To this end, we will proceed as follows. We first review (differentiable) first-order logic and reasoning. We then derive NEMESYS by introducing differentiable logical meta programming. Before concluding, we illustrate several capabilities of NEMESYS.

\section{Background}\label{sec2}

NEMESYS relies on several research areas:
first-order logic, logic programming, differentiable reasoning, meta-reasoning and -learning. 

\textbf{First-Order Logic (FOL)/Logic Programming.} 
A {\it term} is a constant, a variable, or a term which consists of a function symbol. We denote $n$-ary predicate ${\tt p}$ by ${\tt p}/(n,[{\tt dt_1}, \ldots, {\tt dt_n}])$, where ${\tt dt_i}$ is the datatype of $i$-th argument.
An {\it atom} is a formula ${\tt p(t_1, \ldots, t_n) }$, where ${\tt p}$ is an $n$-ary predicate symbol and ${\tt t_1, \ldots, t_n}$ are terms.
A {\it ground atom} or simply a {\it fact} is an atom with no variables.
A {\it literal} is an atom or its negation.
A {\it positive literal} is an atom. 
A {\it negative literal} is the negation of an atom.
A {\it clause} is a finite disjunction ($\lor$) of literals. 
A {\it ground clause} is a clause with no variables.
A {\it definite clause} is a clause with exactly one positive literal.
If  $A, B_1, \ldots, B_n$ are atoms, then $ A \lor \lnot B_1 \lor \ldots \lor \lnot B_n$ is a definite clause.
We write definite clauses in the form of $A~\mbox{:-}~B_1,\ldots,B_n$.
Atom $A$ is called the {\it head}, and a set of negative atoms $\{B_1, \ldots, B_n\}$ is called the {\it body}.
We call definite clauses as clauses for simplicity in this paper.

\textbf{Differentiable Forward-Chaining Reasoning.} The forward-chaining inference is a type of inference in first-order logic to compute logical entailment~\citep{Russel09}. The differentiable forward-chaining inference~\citep{Evans18,Shindo21aaai} computes the logical entailment in a differentiable manner using tensor-based operations. Many extensions of differentiable forward reasoners have been developed, \eg reinforcement learning agents using logic to compute the policy function~\citep{Jiang19logicrl,delfosse2023interpretable} and differentiable rule learners in complex visual scenes~\citep{shindo2023alphailp}. NEMESYS performs differentiable meta-level logic programming based on differentiable forward reasoners.
\begin{figure*}[t]
    \centering
     \includegraphics[width=\textwidth]{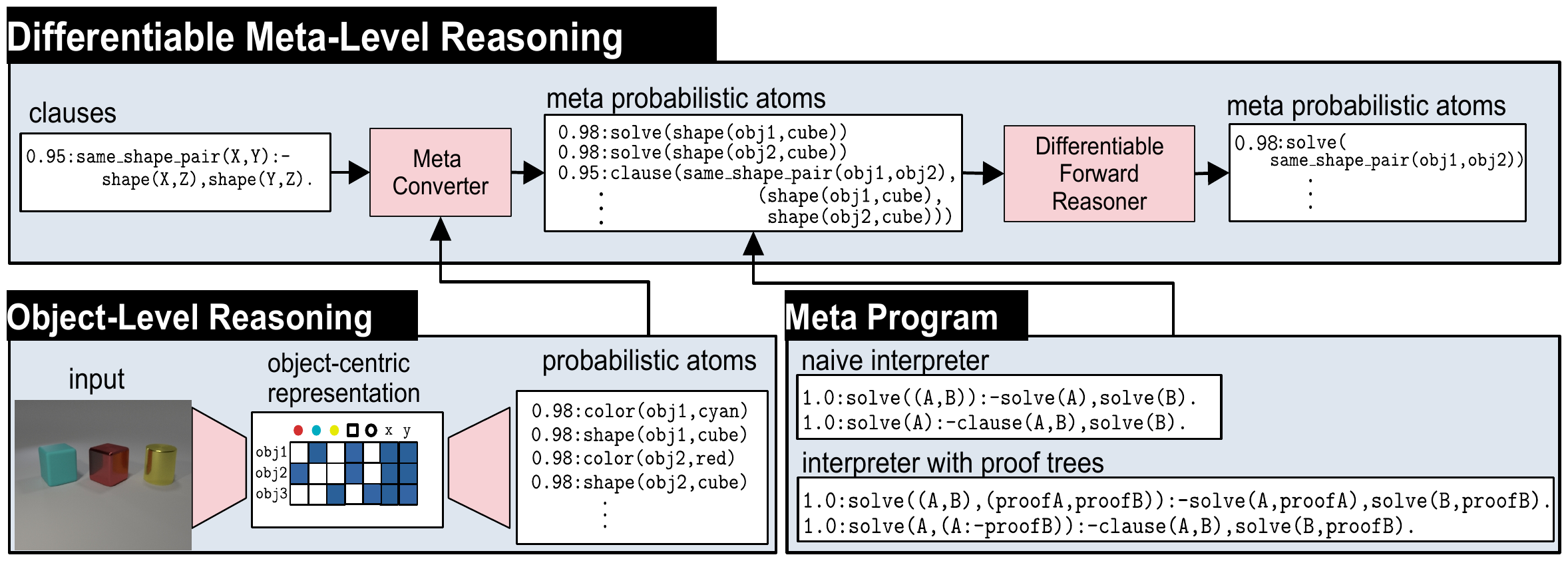}
    \caption{\textbf{Overview of NEMESYS}
    together with an object-level reasoning layer (bottom left). The meta-level reasoner (top) takes a logic program as input, here clauses on the left-hand side in the meta-level reasoning pipeline. Using the meta program (bottom right) it can realize the standard Prolog engine (naive interpreter) or an interpreter that provides \eg~also the proof trees (interpreter with proof trees) without requiring any alterations to the original logic program and internal reasoning function. This means that NEMESYS can integrate many useful functionalities by simply changing devised meta programs without intervening the internal reasoning function. (Best viewed in color) 
    }
    \label{fig:currentpipeline}
\end{figure*}

\textbf{Meta Reasoning and Learning.}
Meta-reasoning is the study about systems which are able to reason about its operation, \ie a system capable of meta-reasoning may be able to reflect, or introspect~\citep{Maes1998Introspection}, shifting from meta-reasoning to object-level reasoning and vice versa~\citep{Costantini2002meta,Griffiths2019meta}.
Compared with imperative programming, it is relatively easier to construct a meta-interpreter using declarative programming. 
First-order Logic~\citep{Lloyd84foundation} has been the major tool to realize  the meta-reasoning systems~\citep{Hill1998metalogic,Pettorossi1992metaproglogic,Apt1995metalogic}. 
For example, Prolog~\citep{sterling1994artofprolog} provides very efficient implementations of meta-interpreters realizing different additional features to the language.

Despite early interest in meta-reasoning within classical Inductive Logic Programming (ILP) systems~\citep{muggleton2014meta,muggleton2014metabayes,muggleton2015meta}, meta-interpreters have remained unexplored within neuro-symbolic AI. 
Meta-interpreters within classical logic are difficult to be combined with gradient-based machine learning paradigms, \emph{e.g.,} deep neural networks.
NEMESYS realizes meta-level reasoning using differentiable forward reasoners in first-order logic, which are able to perform differentiable rule learning on complex visual scenes with deep neural networks~\citep{shindo2023alphailp}. 
Moreover, NEMESYS paves the way to integrate meta-level reasoning into other neuro-symbolic frameworks, including DeepProbLog~\citep{manhaeve2018deepproblog}, Scallop~\citep{huang2021scallop} and NeurASP~\citep{Yang2020neuroasp}, which are rather developed for training neural networks given logic programs using differentiable backward reasoning or answer set semantics. 
We compare NEMESYS with several popular neuro-symbolic systems in three aspects: whether the system
performs \textbf{meta reasoning}, can the same system \textbf{adapt to solve different tasks} and is the
system capable of differentiable \textbf{meta level structure learning}. The comparison results are summarized in Table~\ref{tab:compare_tasks}.


\section{Neural Meta-Symbolic~Reasoning~\&~Learning}\label{sec3}
We now introduce NEMESYS, the first neural meta-symbolic reasoning and learning framework. Fig.~\ref{fig:currentpipeline} shows an overview of NEMESYS.



\subsection{Meta Logic Programming} 
\label{meta}
We describe how meta-level programs are used in the NEMESYS workflow.
In Fig.~\ref{fig:currentpipeline}, the following object-level clause is given as its input:
\begin{align*}
\mathtt{\color{OliveGreen}same\_shape\_pair(X,Y)\color{black} \texttt{:-}\color{Orchid}shape(X,Z),shape(Y,Z)\color{black}.}
\label{eq:same_shape_pair_rule}
\end{align*}
which identifies pairs of objects that have the same shape.
The clause is subsequently fed to \emph{Meta Converter}, which generates meta-level atoms.
Using meta predicate $\mathtt{clause}/2$, the following atom is generated:
\begin{align*}
\mathtt{clause(\color{OliveGreen}same\_shape\_pair(X,Y),} \color{Orchid}\mathtt{(shape(X,Z),shape(Y,Z))\color{black}).}
\end{align*}
where the meta atom $\mathtt{clause(H,B)}$ represents the object-level clause: $\mathtt{H\texttt{:-}B}$.
To perform meta-level reasoning, NEMESYS uses meta-level programs, which often refer to \emph{meta interpreters}, \ie interpreters written in the language itself, as illustrated in Fig.~\ref{fig:currentpipeline}.
For example, a naive interpreter, $\texttt{NaiveInterpreter}$, is defined as:
\begin{align*}
&\mathtt{solve(true)}. \\ 
&\mathtt{solve((A,B))} \mathtt{\texttt{:-} solve(A), solve(B).}\\
&\mathtt{solve(A)} \mathtt{\texttt{:-} clause(\color{OliveGreen}{A},\color{Orchid}{B}\color{black}{), solve(B).}}
\end{align*}
To solve a compound goal $\mathtt{(A, B)}$, we need first solve $\mathtt{A}$ and then $\mathtt{B}$. A single goal $\mathtt{A}$ is solved if there is a clause that rewrites the goal to the new goal $\mathtt{B}$, the body of the clause: $\mathtt{\color{OliveGreen}{A}\texttt{:-}\color{Orchid}{B}}$. This process stops for facts, encoded as $\mathtt{clause(fact,true)}$, since then, $\mathtt{solve(true)}$ will be true.

NEMESYS can employ more enriched meta programs with useful functions by simply changing the meta programs, without modifying the internal reasoning function, as illustrated in the bottom right of Fig.~\ref{fig:currentpipeline}. 
$\texttt{ProofTreeInterpreter}$, an interpreter that produces proof trees along with reasoning, is defined as:
\begin{align*}
&\mathtt{solve(A,(A\texttt{:-}true)).}\\
&\mathtt{solve((A,B),}\mathtt{(proofA,proofB))} \texttt{:-}\mathtt{solve(A,proofA),solve(B,proofB).}\\
&\mathtt{solve(A,}\mathtt{(A\texttt{:-}proofB))} \texttt{:-}\mathtt{clause(\color{OliveGreen}{A},\color{Orchid}{B}\color{black}{), solve(B,proofB).}}\nonumber 
\end{align*}
where $\mathtt{solve(A,Proof)}$ checks if atom $\mathtt{A}$ is true with proof tree $\mathtt{Proof}$.
Using this meta-program, NEMESYS can perform reasoning with integrated proof tree generation.

Now, let us devise the differentiable meta-level reasoning pipeline, which enables NEMESYS to reason and learn flexibly.

\subsection{Differentiable Meta Programming} \label{section:differentiability}

NEMESYS employs differentiable forward reasoning~\cite{shindo2023alphailp}, which computes logical entailment using tensor operations in a differentiable manner, by adapting it to the meta-level atoms and clauses. 

We define a meta-level reasoning function $f^\mathit{reason}_{(\mathcal{C},\mathbf{W})}: [0,1]^G \rightarrow [0,1]^{G}$ parameterized by meta-rules $\mathcal{C}$ and their weights $\mathbf{W}$.
We denote the set of meta-rules by $\mathcal{C}$, and the set of all of the meta-ground atoms by $\mathcal{G}$. $\mathcal{G}$ contains all of the meta-ground atoms produced by a given FOL language.
We consider ordered sets here, \ie each element has its index.
We denote the size of the sets as: $G = |\mathcal{G}|$ and $C = |\mathcal{C}|$.
We denote the $i$-th element of vector $\mathbf{x}$ by $\mathbf{x}[i]$, and the $(i,j)$-th element of matrix $\mathbf{X}$ by $\mathbf{X}[i,j]$. 

First, NEMESYS converts visual input to a \emph{valuation vector} $\mathbf{v} \in [0,1]^G$, which maps each meta atom to a probabilistic value (Fig.~\ref{fig:currentpipeline} \emph{Meta Converter}).
For example,
\begin{equation*}
  \mathbf{v}=
  \begin{blockarray}{cl}
    \begin{block}{[c]l}
      0.98  & \mathtt{solve(color(obj1,\ cyan))} \\
      0.01 & \mathtt{solve(color(obj1,\ red))} \\
      0.95  & \mathtt{clause(same\_shape\_pair(\ldots),\ (shape(\ldots),\ \ldots))} \\
    \end{block}
  \end{blockarray}
  \vspace{-1.2em}
\end{equation*}
represents a valuation vector that maps each meta-ground atom to a probabilistic value. For readability, only selected atoms are shown.
NEMESYS computes logical entailment by updating the initial valuation vector $\mathbf{v}^{(0)}$ for $T$ times to $\mathbf{v}^{(T)}$.



Subsequently, we compose the reasoning function that computes logical entailment.
We now describe each step in detail.

\paragraph{(Step 1) Encode Logic Programs to Tensors.} 
To achieve differentiable forward reasoning, each meta-rule is encoded to a tensor representation. 
Let $S$ be the maximum number of substitutions for existentially quantified variables in $\mathcal{C}$, and $L$ be the maximum length of the body of rules in $\mathcal{C}$.
Each meta-rule $C_i \in \mathcal{C}$ is encoded to a tensor ${\bf I}_i \in \mathbb{N}^{G \times S \times L}$, which contains the indices of body atoms. Intuitively, $\mathbf{I}_i[j,k,l]$ is the index of the $l$-th fact (subgoal) in the body of the $i$-th rule to derive the $j$-th fact with the $k$-th substitution for existentially quantified variables. We obtain $\mathbf{I}_i$ by firstly grounding the meta rule $C_i$, then computing the indices of the ground body atoms, and transforming them into a tensor.

\paragraph{(Step 2) Assign Meta-Rule Weights.}
We assign weights to compose the reasoning function with several meta-rules as follows:
(i) We fix the target programs' size as $M$, \ie we try to select a meta-program with $M$ meta-rules out of $C$ candidate meta rules.
(ii)  We introduce $C$-dimensional weights $\mathbf{W} = [ {\bf w}_1, \ldots, {\bf w}_M ]$ where $\mathbf{w}_i \in \mathbb{R}^C$.
(iii) We take the \emph{softmax} of each weight vector ${\bf w}_j \in \mathbf{W}$ and softly choose $M$ meta rules out of $C$ meta rules to compose the differentiable meta program.

\paragraph{(Step 3) Perform Differentiable Inference.}
We compute $1$-step forward reasoning using weighted meta-rules, then we recursively perform reasoning to compute $T$-step reasoning.

\textbf{(i) Reasoning using one rule.} First, for each meta-rule $C_i \in \mathcal{C}$, we evaluate body atoms for different grounding of $C_i$ by computing:
\begin{align}
   b_{i,j,k}^{(t)} = \prod_{1 \leq l \leq L} {\bf gather}({\bf v}^{(t)}, {\bf I}_i)[j,k,l],
   \label{eq:gather_prod_body}
\end{align}
where $\mathbf{gather}: [0,1]^{G} \times \mathbb{N}^{G \times S \times L} \rightarrow [0,1]^{G \times S \times L}$ is:
\begin{align}
    \mathbf{gather}({\bf x}, {\bf Y})[j,k,l] = {\bf x}[{\bf Y}[j,k,l]], 
\end{align}
and $b^{(t)}_{i,j,k} \in [0,1]$. The $\mathbf{gather}$ function replaces the indices of the body atoms by the current valuation values in $\mathbf{v}^{(t)}$.
To take logical {\em and} across the subgoals in the body, we take the product across valuations.
$b_{i,j,k}^{(t)}$ represents the valuation of body atoms for $i$-th meta-rule using $k$-th substitution for the existentially quantified variables to deduce $j$-th meta-ground atom at time $t$.

Now we take logical {\em or} softly to combine all of the different grounding for $C_i$ by computing $c^{(t)}_{i,j} \in [0,1]$:
\begin{align}
    c^{(t)}_{i,j} = \mathit{softor}^\gamma (b_{i,j,1}^{(t)}, \ldots, b_{i,j,S}^{(t)}),
\end{align}
where $\mathit{softor}^\gamma$ is a smooth logical {\em or}  function: 
\begin{align}
    \mathit{softor}^\gamma(x_1, \ldots, x_n) = \gamma \log \sum_{1\leq i \leq n} \exp(x_i / \gamma),
    \label{eq:softor}
\end{align}
where $\gamma > 0$ is a smooth parameter.
Eq.~\ref{eq:softor} is an approximation of the \emph{max} function over probabilistic values based on the \emph{log-sum-exp} approach~\citep{cuturi2017soft}.

\textbf{(ii) Combine results from different rules.}
Now we apply different meta-rules using the assigned weights by computing:
\begin{align}
     h_{j,m}^{(t)} = \sum_{1 \leq i \leq C}  w^*_{m,i} \cdot c_{i,j}^{(t)},
\end{align}
where $h_{j,m}^{(t)} \in [0,1]$, $w^*_{m,i} = \exp(w_{m,i})/{\sum_{i^\prime} \exp(w_{m, i^\prime})}$, 
and $w_{m,i} = \mathbf{w}_m[i]$.
Note that $w^*_{m,i}$ is interpreted as a probability that meta-rule $C_i \in \mathcal{C}$ is the $m$-th component.
We complete the $1$-step forward reasoning by combining the results from different weights:
\begin{align}
    r_{j}^{(t)} = \mathit{softor}^\gamma ( h_{j,1}^{(t)}, \ldots, h_{j,M}^{(t)} ).
    \label{eq:softor_weighted_rules}
\end{align}
Taking $\mathit{softor}^\gamma$ means that we compose $M$ softly chosen rules out of $C$ candidate meta-rules.

\textbf{(iii) Multi-step reasoning.} We perform $T$-step forward reasoning by computing $r_j^{(t)}$ recursively for $T$ times:
$v^{(t+1)}_j = \mathit{softor}^\gamma (r^{(t)}_j, v^{(t)}_j)$.
Updating the valuation vector for $T$-times corresponds to computing logical entailment softly by $T$-step forward reasoning.
The whole reasoning computation Eq.~\ref{eq:gather_prod_body}-\ref{eq:softor_weighted_rules} can be implemented using efficient tensor operations.

\section{Experiments}
\label{sec4}
With the methodology of NEMESYS established, we subsequently provide empirical evidence of its benefits over neural baselines and object-level neuro-symbolic approaches:
\textbf{(1)} NEMESYS can emulate a differentiable forward reasoner, \ie it is a sufficient implementation of object-centric reasoners with a naive meta program.
\textbf{(2)} NEMESYS is capable of differentiable meta-level reasoning, \ie it can integrate additional useful functions using devised meta-rules. We demonstrate this advantage by solving tasks of proof-tree generation, relevance propagation, automated planning, and causal reasoning. 
\textbf{(3)} NEMESYS can perform parameter and structure learning 
efficiently using gradient descent, \ie it can perform learning on meta-level programs. 


In our experiments, we implemented NEMESYS in Python using PyTorch, with CPU: intel i7-10750H and RAM: 16 GB.
 \begin{table*}[t]
\small
\centering
\begin{tabular}{lrrr}
      & \textbf{NEMESYS} & \textbf{ResNet50} & \textbf{YOLO+MLP} \\ 
Twopairs     & 100.0$\bullet$                    & 50.81\hspace{1ex}                       & 98.07$\circ$    \\
Threepairs   & 100.0$\bullet$                       &51.65\hspace{1ex}                               & 91.27$\circ$       \\
Closeby      & 100.0$\bullet$                        & 54.53\hspace{1ex}                           & 91.40$\circ$    \\
Red-Triangle & 95.6$\bullet$                       & 57.19\hspace{1ex}                           & 78.37$\circ$        \\
Online/Pair  & 100.0$\bullet$                      & 51.86\hspace{1ex}                          & 66.19$\circ$      \\
9-Circles    & 95.2$\bullet$                      & 50.76$\circ$                         & 50.76$\circ$     
\end{tabular}
\caption{\textbf{Performance (accuracy; the higher, the better)} on the test split of Kandinsky patterns. The best-performing models are denoted using $\bullet$, and the runner-up using  $\circ$. In Kandinsky patterns, NEMESYS produced almost perfect accuracies outperforming neural baselines, where \textbf{YOLO+MLP} is a neural baseline using pre-trained YOLO~\citep{YOLO} combined with a simple MLP, showing the capability of solving complex visual reasoning tasks. The performances of baselines are shown in \citep{Shindo21nsfr}.} 
\label{table:experiment1}
\end{table*}
\subsection{Visual Reasoning on Complex Pattenrs}
\label{dvrl}
Let us start off by showing that our NEMESYS is able to obtain the equivalent high-quality results as a standard object-level reasoner but on the meta-level. 
We considered tasks of Kandinsky patterns~\citep{Holzinger19IQ,Holzinger21} and CLEVR-Hans~\citep{Stammer21clevrhans}\footnote{We refer to \cite{Stammer21clevrhans} and \citep{Shindo21nsfr} for detailed explanations of the used patterns for CLEVR-Hans and Kandinsky patterns.}. 
CLEVR-Hans is a classification task of complex 3D visual scenes.
We compared NEMESYS with the naive interpreter against neural baselines and a neuro-symbolic baseline, $\alpha$ILP~\citep{shindo2023alphailp}, which achieves state-of-the-art performance on these tasks. 
For all tasks, NEMESYS achieved exactly the same performances with $\alpha$ILP since the naive interpreter realizes a conventional object-centric reasoner.
Moreover, as shown in Table~\ref{table:experiment1} and Table~\ref{table:experiment2}, NEMESYS outperformed neural baselines on each task. This shows that NEMESYS is able to solve complex visual reasoning tasks using meta-level reasoning without sacrificing performance.

In contrast to the object-centric reasoners, \eg $\alpha$ILP. NEMESYS can easily integrate additional useful functions by simply switching or adding meta programs without modifying the internal reasoning function, as shown in the next experiments. 
\subsection{Explainable Logical Reasoning}
One of the major limitations of differentiable forward chaining~\citep{Evans18,Shindo21aaai,shindo2023alphailp} is that they lack the ability to \emph{explain} the reasoning steps and their evidence.
We show that NEMESYS achieves explainable reasoning by incorporating devised meta-level programs.

\paragraph{Reasoning with Integrated Proof Tree Generation} 
\label{prooftree}
First, we demonstrate that NEMESYS can generate proof trees while performing reasoning, which the previous differentiable forward reasoners cannot produce since they encode the reasoning function to computational graphs using tensor operations and observe only their input and output.
Since NEMESYS performs reasoning using meta-level programs, it can add the function to produce proof trees into its underlying reasoning mechanism simply by devising them, as illustrated in Fig~\ref{fig:currentpipeline}. 

\begin{figure*}[t]
    \centering
    \includegraphics[trim=0 335 0 90, clip,width=\textwidth]{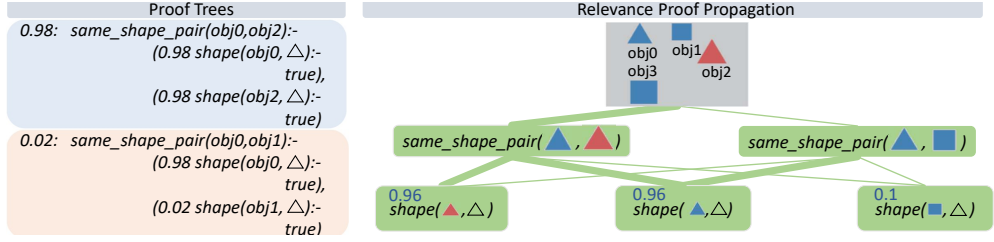}
    \caption{
    \textbf{NEMESYS explains its reasoning with proof trees and relevance proof propagation. }
    Given the image involving four objects (top, right), NEMESYS provides two proofs (two boxes on the left, true atom's proof (blue box) and false atom's proof (cream box)). They can be leveraged to decompose the prediction of NEMESYS into relevance scores per (ground) atom (right). 
    First, a standard forward reasoning is performed to compute the prediction. Then, the model's prediction is backward propagated through the proof trees by applying specific decomposition rules, see main text.
    The numbers next to each (ground) atom are the relevance scores computed. The larger the score is, the more impact an (ground) atom has on the final prediction, and the line width is wider. For brevity, the complete proof tree is not depicted here. As our baseline comparison, we extend DeepProbLog~\citep{manhaeve2018deepproblog} to DeepMetaProbLog. However, DeepMetaProbLog only provides proof tree for true atoms (top left blue box). (Best viewed in color)}
    \label{fig:prooftreene}
\end{figure*}

We use Kandinsky patterns~\cite{Mueller21}, a benchmark of visual reasoning whose classification rule is defined on high-level concepts of relations and attributes of objects. We illustrate the input on the top right of Fig.~\ref{fig:prooftreene} that belongs to a pattern: \emph{"There are two pairs of objects that share the same shape."}
Given the visual input, proof trees generated using the $\texttt{ProofTreeInterpreter}$ in Sec.~\ref{meta} are shown on the left two boxes of Fig.~\ref{fig:prooftreene}. 
In this experiment, NEMESYS identified the relations between objects, and the generated proof trees explain the intermediate reasoning steps. 



\begin{table*}[t]
\small
\centering
\begin{tabular}{lrrrr}
                                & \multicolumn{2}{c}{\bf CLEVR-Hans3}                                       & \multicolumn{2}{c}{\bf CLEVR-Hans7}                            \\ 
        & \multicolumn{1}{c}{Validation} & \multicolumn{1}{c}{Test}           & \multicolumn{1}{l}{Validation} & \multicolumn{1}{l}{Test} \\ 
    
CNN            & 99.55$\circ$ & \multicolumn{1}{r|}{70.34 \ \hspace{1ex}}          & 96.09\hspace{1ex}                           & 84.50\hspace{1ex}        \\
NeSy (Default) & 98.55\hspace{1ex}                           & \multicolumn{1}{r|}{81.71 \ \hspace{1ex}}          & 96.88$\circ$                           & 90.97\hspace{1ex}                    \\
NeSy-XIL       & 100.00$\bullet$                          & \multicolumn{1}{r|}{91.31$\circ$}          & 98.76$\bullet$                           & 94.96$\bullet$           \\ 
NEMESYS          & 98.18\hspace{1ex}                           & \multicolumn{1}{c|}{98.40$\bullet$} & 93.60\hspace{1ex}                           & 92.19$\circ$                   
\end{tabular}
\caption{\textbf{Performance (accuracy; the higher, the better)} on the validation/test splits of 3D CLEVR-Hans data sets. The best-performing models are denoted using $\bullet$, and the runner-up using  $\circ$. In CLEVR-Hans, NEMESYS outperformed neural baselines including: \textbf{(CNN)} A ResNet~\citep{He_2016_Residualnetwork}, \textbf{(NeSy)} A model combining object-centric model (Slot Attention~\citep{slotattention} and Set Transformer~\citep{Lee19settransformer}, and \textbf{(NeSy-XIL)} Slot Attention and Set Transformer using human feedback. NEMESYS tends to show less overfitting and performs similarly to a neuro-symbolic approach using human feedback (NeSy-XIL). The performances of baselines are shown in \citep{Stammer21clevrhans} and \cite{Shindo21nsfr}.}
\label{table:experiment2}
\vspace{-0.1in}
\end{table*}

Let's first consider the top left blue box depicted in Fig.~\ref{fig:prooftreene} (for readability, we only show the proof part of meta atoms in the image). The weighted ground atom $\mathtt{0.98:}\mathtt{same\_shape\_pair(obj0,obj2)}$ proves $\mathtt{obj0}$ and $\mathtt{obj2}$ are of the same shape with the probability $0.98$. The proof part shows that NEMESYS comes to this conclusion since both objects are triangle with probabilities of $\mathtt{0.98}$ and in turn it can apply the rule for $\mathtt{same\_shape\_pair}$. 
We use this example to show how to compute the weight of the meta atoms inside NEMESYS. With the proof-tree meta rules and corresponding meta ground atoms:
\begin{align*}
&\mathtt{0.98:}\ \color{Plum}\mathtt{solve(shape(obj0,} \text{\includegraphics[height=1.5ex]{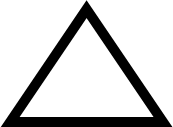}}\mathtt{),(shape(obj0,} \text{\includegraphics[height=1.5ex]{plots/triangle.png}}\mathtt{),true).}  \\
&\mathtt{0.98:}\ \color{BlueViolet}\mathtt{solve(shape(obj1,} \text{\includegraphics[height=1.5ex]{plots/triangle.png}}\mathtt{),(shape(obj1,} \text{\includegraphics[height=1.5ex]{plots/triangle.png}}\mathtt{),true).} \\
&\color{OliveGreen}\mathtt{solve((A,B),(proofA,proofB))} \texttt{:-} \color{BlueViolet}\mathtt{solve(A,proofA),} \color{Plum}\mathtt{solve(B,proofB).}\\
&\color{Orchid}\mathtt{solve(A,(A\texttt{:-}proofB))}\color{Black} \texttt{:-} \color{CadetBlue}\mathtt{clause(\{A\},\{B\}),} \color{OliveGreen}\mathtt{solve(B,proofB).}
\end{align*}
The weight of the meta ground atoms are computed by \emph{Meta Converter} when mapping the probability of meta ground atoms to a continuous value. The meta ground atom says that $\mathtt{shape(obj0,\includegraphics[height=1.5ex]{plots/triangle.png})}$ is true with a high probability of $0.98$ because $\mathtt{shape(obj0,\includegraphics[height=1.5ex]{plots/triangle.png})}$ can be proven. 

With the two meta ground atoms at hand, we infer the weight of the meta atom with compound goals $\color{OliveGreen}\mathtt{solve((shape(obj0,\includegraphics[height=1.5ex]{plots/triangle.png}),shape(obj2,\includegraphics[height=1.5ex]{plots/triangle.png})),(ProofA, ProofB))}$\color{Black}, based on the first meta rule (for readability, we omit writing out the proof part). Then, we use the second meta rule to compute the weight of the meta atom $
\color{Orchid}\mathtt{solve}\mathtt{(same\_shape\_pair(obj0,obj2)},\mathtt{(Proof))}$ \color{Black}, using the compound goal meta atom $\color{OliveGreen}\mathtt{solve((shape(obj0,\includegraphics[height=1.5ex]{plots/triangle.png}),shape(obj2,\includegraphics[height=1.5ex]{plots/triangle.png})),(ProofA, ProofB))}$\color{Black} and the meta atom \color{CadetBlue}
$
\mathtt{clause}\mathtt{(same\_shape\_pair(obj0,obj2)},\mathtt{(shape(obj0,\includegraphics[height=1.5ex]{plots/triangle.png})},\mathtt{shape(obj2,\includegraphics[height=1.5ex]{plots/triangle.png})))}
$ \color{Black}.

In contrast, NEMESYS can explicitly show that $\mathtt{obj0}$ and $\mathtt{obj1}$ have a low probability of being of the same shape (Fig.~\ref{fig:prooftreene} left bottom cream box).
This proof tree shows that the goal $\mathtt{shape(obj1,\includegraphics[height=1.5ex]{plots/triangle.png})}$ has a low probability of being true. Thus, as one can read off, $\mathtt{obj0}$ is most likely a triangle, while $\mathtt{obj1}$ is most likely not a triangle. In turn, NEMESYS concludes with a low probability that $\mathtt{same\_shape\_pair(obj0,obj1)}$ is true, only a probability of $0.02$. 
NEMESYS can produce all the information required to explain its decisions by simply changing the meta-program, not the underlying reasoning system.

\textbf{Using meta programming to extend DeepProbLog to produce proof trees as a baseline comparison}. Since DeepProbLog~\citep{manhaeve2018deepproblog} doesn't support generating proof trees in parallel with reasoning, 
we extend DeepProbLog~\citep{manhaeve2018deepproblog} to DeepMetaProblog to generate proof trees as our baseline comparison using ProbLog~\citep{de2007problog}. 
However, the proof tree generated by DeepMetaProbLog is limited to the `true' atoms (Fig.~\ref{fig:prooftreene} top left blue box), \ie DeepMetaProbLog is unable to generate proof tree for false atoms such as $\mathtt{same\_shape\_pair(obj0,obj1)}$ (Fig.~\ref{fig:prooftreene} bottom left cream box) due to backward reasoning.

\paragraph{Logical Relevance Proof Propagation (LRP$^2$)} 
Inspired by layer-wise relevance propagation (LRP)~\citep{lapuschkin2019unmasking}, which produces explanations for feed-forward neural networks,  we now show that, LRP can be adapted to logical reasoning systems using declarative languages in NEMESYS, thereby enabling the reasoning system to articulate the rationale behind its decisions, 
\ie it can compute the importance of ground atoms for a query by having access to proof trees. We call this process: \emph{logical relevance proof propagation} (LRP$^2$). 

The original LRP technique decomposes the prediction of the network, $f(\mathbf{x})$, onto the input variables, $\mathbf{x}=\left(x_1, \ldots, x_d\right)$, through a decomposition $\mathbf{R}=\left(R_1, \ldots, R_d\right)$ such that $\sum\nolimits_{p=1}^d R_p= f(\mathbf{x})\;$. Given the activation  
$a_j=\rho\left(\sum_i a_i w_{ij}+b_j\right)$ of neuron, where $i$ and $j$ denote the neuron indices at consecutive layers, and $\sum_i$ and $\sum_j$ represent the summation over all neurons in the respective layers, the propagation of LRP is defined as: $R_i=\sum\nolimits_j z_{i j} ({\sum\nolimits_i z_{i j}})^{-1} R_j,$ where $z_{ij}$ is the contribution of neuron $i$ to the activation $a_j$, typically some function of activation $a_i$ and the weight $w_{ij}$. Starting from the output $f(\mathbf{x})$, the relevance is computed layer by layer until the input variables are reached.
 

To adapt this in NEMESYS to ground atoms and proof trees, we have to be a bit careful, since we cannot deal with the uncountable, infinite real numbers within our logic. Fortunately, we can make use of the weight associated with ground atoms.
That is, our LRP$^2$ composes meta-level atoms that represent the relevance of an atom given proof trees and associates the relevance scores to the weights of the meta-level atoms. 

To this end, we introduce \emph{three} meta predicates:
$\mathtt{rp/3/[goal,proofs,atom]}$ that represents the relevance score an $\mathtt{atom}$ has on the $\mathtt{goal}$ in given $\mathtt{proofs}$, 
$\mathtt{assert\_probs/1/[atom]}$ that looks up the valuations of the ground atoms and maps the probability of the $\mathtt{atom}$ to its weight. $\mathtt{rpf/2/[proof, atom]}$ represents how much an $\mathtt{atom}$ contributes to the $\mathtt{proof}$. The atom $\mathtt{assert\_probs((Goal\texttt{:-}Body))}$ asserts the probability of the atom $\mathtt{(Goal\texttt{:-}Body)}$. 
With them, the meta-level program of LRP$^2$ is:
\begin{align*}
&\mathtt{rp(Goal,Body,Atom)}\texttt{:-}\mathtt{assert\_probs((Goal\texttt{:-}Body))},\\ &\hspace{23ex}\mathtt{assert\_probs(Body)},\mathtt{rpf(Body,Atom)}.\\
&\mathtt{rpf([Atom,Tail],Atom)}\texttt{:-}\mathtt{assert\_probs(Atom)}.\\
&\mathtt{rpf([Head,Tail],Atom)}\texttt{:-}\mathtt{rpf(Tail,Atom)}.\\
&\mathtt{rpf([Head,Tail],Atom)}\texttt{:-}\mathtt{clause(Head,Body)},\mathtt{rpf(Body,Atom)}.\\
&\mathtt{rpf([Head,Tail],Atom)}\texttt{:-}\mathtt{clause(Head,Body)},\mathtt{rpf(Tail,Atom)}.\\
&\mathtt{rpf(Atom,Atom)}\texttt{:-}\mathtt{assert\_probs(Atom)}.\\
&\mathtt{rpf(Atom\_B,Atom)}\texttt{:-}\mathtt{norelate(Atom\_B,Atom)}.\\
&\mathtt{rp(Goal,[Proof,Proofs],Atom)}\texttt{:-}\mathtt{rp[Goal,[Proof],Atom]}.\\
&\mathtt{rp(Goal,[Proof,Proofs],Atom)}\texttt{:-}\mathtt{rp[Goal,Proofs,Atom]}.
\end{align*}
where $\mathtt{rp(Goal,Proof,Atom)}$ represents the relevance score an $\mathtt{Atom}$ has on the $\mathtt{Goal}$ in a $\mathtt{Proof}$, \ie 
we interpret the associated weight with atom $\mathtt{rp(Goal,Proof,Atom)}$ as the actual relevance score of $\mathtt{Atom}$ has on $\mathtt{Goal}$ given $\mathtt{Proof}$.
The higher the weight of $\mathtt{rp(Goal,Proof,Atom)}$ is, the larger the impact of $\mathtt{Atom}$ has on the $\mathtt{Goal}$. 

Let us go through the meta rules of LRP$^2$. 
The first rule defines how to compute the relevance score of an $\mathtt{Atom}$ given the $\mathtt{Goal}$ under the condition of a $\mathtt{Body}$ (a single $\mathtt{Proof}$). The relevance score is computed by multiplying the weight of the $\mathtt{Body}$, the weight of a clause $\mathtt{(Goal\texttt{:-}Body)}$ and the importance score of the $\mathtt{Atom}$ given the $\mathtt{Body}$. 
The second to the seventh rule defines how to calculate the importance score of an $\mathtt{Atom}$ given a $\mathtt{Proof}$. These six rules loop over each atom of the given $\mathtt{Proof}$, once it detects the $\mathtt{Atom}$ inside the given $\mathtt{Proof}$, the importance score will be set to the weight of the $\mathtt{Atom}$, another case is that the $\mathtt{Atom}$ is not in $\mathtt{Proof}$, in that case, in the seventh rule, $\mathtt{norelate}$ will set the importance score to a small value.
The eighth and ninth rules amalgamate the results from different proofs, \ie the score from each proof tree is computed recursively during forward reasoning. The scores for the same target (the pair of $\mathtt{Atom}$ and $\mathtt{Goal}$) are combined by the $\mathit{softor}$ operation. The score of an atom given several proofs is computed by taking \emph{logical or} softly over scores from each proof.

With these \emph{nine} meta rules at hand, together with the proof tree, NEMESYS is able to perform the relevance proof propagation for different atoms. We consider using the proof tree generated in Sec.~\ref{prooftree} and set the goal as: $\mathtt{same\_shape\_pair(obj0,obj2)}$. 
Fig.~\ref{fig:prooftreene} (right) shows LRP$^2$-based explanations generated by NEMESYS.
The relevance scores of different ground atoms are listed next to each (ground) atom. As we can see, the atoms $\mathtt{shape(obj0,\includegraphics[height=1.5ex]{plots/triangle.png})}$ and $\mathtt{shape(obj2,\includegraphics[height=1.5ex]{plots/triangle.png})}$ have the largest impact on the goal $\mathtt{same\_shape\_pair(obj0,obj2)}$, while $\mathtt{shape(obj1,\includegraphics[height=1.5ex]{plots/triangle.png})}$ have much smaller impact. 

By providing proof tree and LRP$^2$, NEMESYS computes the precise effect of a ground atom on the goal and produces an accurate proof to support its conclusion. This approach is distinct from the Most Probable Explanation (MPE) \citep{kwisthout2011mpe}, which generates the most probable proof rather than the exact proof.


\begin{figure}[t]
    \centering
    \includegraphics[width=0.243\textwidth]{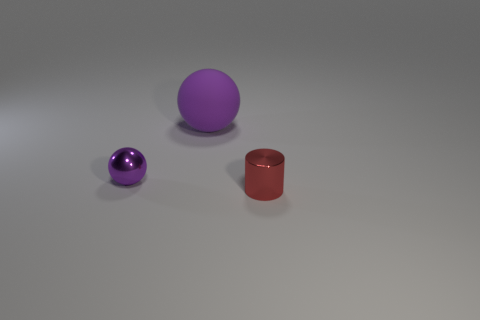}
    \includegraphics[width=0.243\columnwidth]{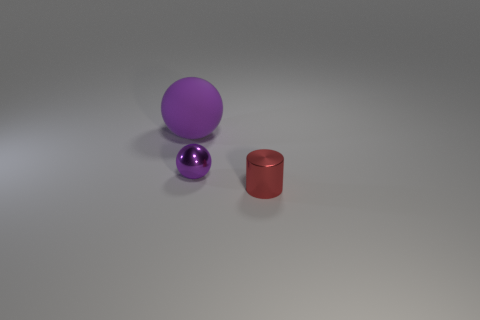}
    \includegraphics[width=0.243\columnwidth]{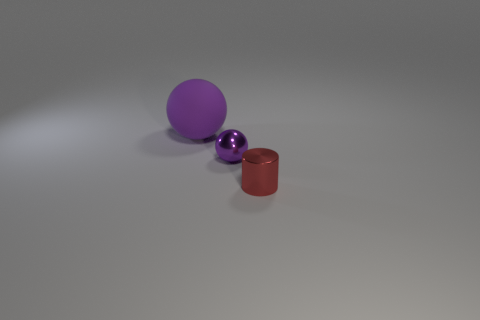} 
    \includegraphics[width=0.243\textwidth]{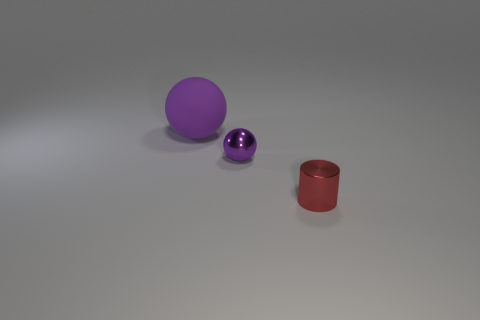}
    \caption{\textbf{Visual Concept Repairing: NEMESYS achieves planning by performing differentiable meta-level reasoning}. The left most image shows the \emph{start} state, and the right most image shows the \emph{goal} state. Taking these states as inputs, NEMESYS performs differentiable forward reasoning using meta-level clauses that simulate the planning steps and generate intermediate states (two images in the middle) and actions from \emph{start} state to reach the \emph{goal} state. (Best viewed in color)}
    \label{fig:clever}
\end{figure}

\subsection{Avoiding Infinite Loops}
 
Differentiable forward chaining~\citep{Shindo21aaai}, unfortunately, can generate infinite computations. A pathological example:
\begin{align*}
    &\mathtt{edge(a,b). \ edge(b,a).} \ \mathtt{edge(b,c).} \quad \mathtt{path(A,A,[\ ]).} \quad \\ 
    &\mathtt{path(A,C,[edge(A,B),}\mathtt{path])}\texttt{:-}\mathtt{edge(A,B),path(B,C,[path]).}
\end{align*}
\begin{wrapfigure}{t}{0.25\textwidth}
     \centering
     \includegraphics[width=0.25\textwidth]{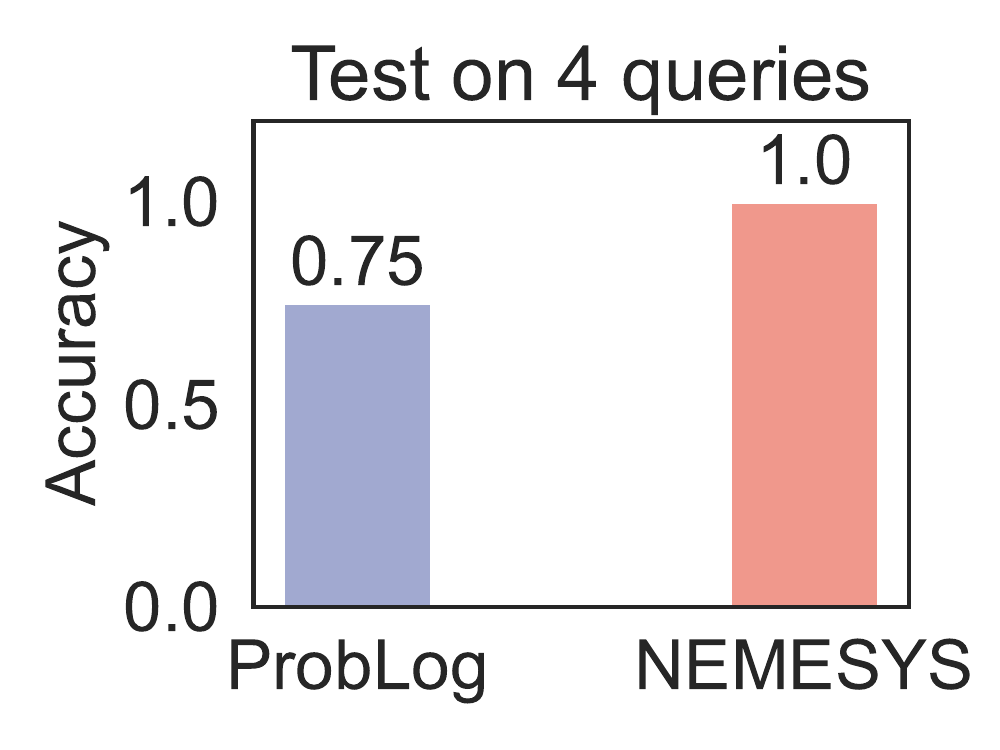}
          \caption{\textbf{Performance (accuracy; the higher, the better)on four queries.} (Best viewed in color)}
     \label{fig:infiniteloop}
\end{wrapfigure}
It defines a simple graph over three nodes $(a, b, c)$ with three edges, $(a-b, b-a, b-c)$ as well as paths in graphs in general. Specifically, $\mathtt{path}/3$ defines how to find a path between two nodes in a recursive way. The base case is $\mathtt{path(A,A,[])}$, meaning that any node $\mathtt{A}$ is reachable from itself. 
The recursion then says, if there is an edge from node $\mathtt{A}$ to node $\mathtt{B}$, and there is a path from node $\mathtt{B}$ to node $\mathtt{C}$, then there is a path from node $\mathtt{A}$ to node $\mathtt{C}$. Unfortunately, this generates an infinite loop $\mathtt{[edge(a,b),edge(b,a),edge(a,b),\ldots]}$ when computing the path from $a$ to $c$, since this path can always be extended potentially also leading to the node $c$.

Fortunately, NEMESYS allows one to avoid infinite loops by memorizing the proof-depth, \ie we simply implement a limited proof-depth strategy on the meta-level: 
\begin{align*}
    &\mathtt{li((A,B), DPT)} \texttt{:-}\mathtt{li(A, DPT)},\mathtt{li(B, DPT).}\\
    &\mathtt{li(A,DPT)} \texttt{:-}\mathtt{clause(A,B),}\ \mathtt{DPT1} \ \texttt{is}\  \mathtt{DPT - 1,} \ \mathtt{li(B, DPT1).}
\end{align*}
With this proof strategy, NEMESYS gets the path $\mathtt{path(a,c,[edge(a,b),edge(b,c)]) = true}$ in three steps. For simplicity, we omit the proof part in the atom. Using the second rule and the first rule recursively, the meta interpreter finds $\mathtt{clause(path(a,c),(edge(a,b),path(b,c)))}$  and $\mathtt{clause(path(b,c),(edge(b,c),path(c,c)))}$. Finally, 
the meta interpreter finds a clause, whose head is $\mathtt{li(path(c,c),1)}$ and the body is $\mbox{\em true}$. 

Since forward chaining gets stuck in the infinite loop, we choose ProbLog~\citep{de2007problog} as our baseline comparison. We test NEMESYS and ProbLog using four queries, including one query which calls the recursive rule. ProbLog fails to return the correct answer on the query which calls the recursive rule. The comparison is summarized in Fig.~\ref{fig:infiniteloop}. We provide the four test queries in Appendix~\ref{loop}. 


\subsection{Differentiable First-Order Logical Planning}

As the fourth meta interpreter, we demonstrate NEMESYS as a differentiable planner. Consider Fig.~\ref{fig:clever} where NEMESYS was asked to put all objects of a start image onto a line. 
Each start and goal state is represented as a visual scene, which is generated in the CLEVR~\citep{Johnson17clevr} environment.
 By adopting a perception model, \eg YOLO~\citep{YOLO} or slot attention~\citep{slotattention}, NEMESYS obtains logical representations of the start and end states: 
\begin{align*}
\mathtt{start} &= \{ \mathtt{pos(obj0, (1, 3)), \ldots, pos(obj4, (2, 1))} \},\\
\mathtt{goal} &= \{ \mathtt{pos(obj0, (1, 1)), \ldots, pos(obj4, (5, 5))} \},
\end{align*}
where $\mathtt{pos/2}$ describes the $2$-dim positions of objects. NEMESYS solves this planning task by performing differentiable reasoning using the meta-level program:
\begin{align*}
&\mathtt{plan(Start\_state,}\mathtt{New\_state,Goal\_state,[Action,Old\_stack])}\textbf{:-}\\
&\hspace{25ex}\mathtt{move(Action,Old\_state,New\_state),}\\
&\hspace{25ex}\mathtt{condition\_met(Old\_state,Current\_state),}\\
&\hspace{25ex}\mathtt{change\_state(Current\_state,New\_state),} \\
&\hspace{25ex}\mathtt{plan(Start\_state,Current\_state,Goal\_state,Old\_stack).}\\
&\mathtt{planf(Start\_state,}\mathtt{Goal,Move\_stack)}\textbf{:-}\\
&\hspace{24ex}\mathtt{plan(Start\_state,Current\_state,Goal\_state,Move\_stack),}\\ &\hspace{24ex}\mathtt{equal(Current\_state,Goal\_state).}
\end{align*}




The first meta rule presents the recursive rule for plan generation, and the second rule gives the successful termination condition for the plan when the $\mathtt{Goal\_state}$ is reached, where  $\mathtt{equal/2}$ checks whether the $\mathtt{Current\_state}$ is the $\mathtt{Goal\_state}$ and the $\mathtt{planf/3}$ contains $\mathtt{Start\_state}$, $\mathtt{Goal\_state}$ and the needed action sequences $\mathtt{Move\_stack}$ from $\mathtt{Start\_state}$ to reach the $\mathtt{Goal\_state}$. 

The predicate $\mathtt{plan/4}$ takes four entries as inputs: $\mathtt{Start\_state}$, $\mathtt{State}$, $\mathtt{Goal\_state}$ and $\mathtt{Move\_stack}$. The $\mathtt{move/3}$ predicate uses $\mathtt{Action}$ to push $\mathtt{Old\_state}$ to $\mathtt{New\_state}$. $\mathtt{condition\_met/2}$ checks if the state’s preconditions are met. When the preconditions are met, $\mathtt{change\_state/2}$ changes the state, and $\mathtt{plan/4}$  continues the recursive search. 

To reduce memory usage, we split the move action in horizontal and vertical in the experiment. 
For example, NEMESYS represents an action to move an object in the horizontal direction right by $\mathtt{1}$ step
using meta-level atom:
\begin{align*}
\mathtt{move(}&\mathtt{move\_right}, \mathtt{pos\_hori(Object,X),}\mathtt{pos\_hori(Object,X}\texttt{+}\mathtt{1)).}
\end{align*}
where $\mathtt{move\_right}$ represents the action,
$\mathtt{X+1}$ represents arithmetic sums over (positive) integers, encoded as $\mathtt{0,succ(0),succ(succ(0))}$ and so on as terms.
Performing reasoning on the meta-level clause with $\mathtt{plan}$ simulates a step as a planner, \ie it computes preconditions, and applies actions to compute states after taking the actions.
Fig.~\ref{fig:clever} summarizes one of the experiments performed using NEMESYS on the Visual Concept Repairing task.
We provided the start and goal states as visual scenes containing varying numbers of objects with different attributes.
The left most image of Fig.~\ref{fig:clever} shows the start state, and the right most image shows the goal state, respectively.
NEMESYS successfully moved objects to form a line.
For example, to move $\mathtt{obj0}$ from $\mathtt{(1,1)}$ to $\mathtt{(3,1)}$, NEMESYS deduces:
\begin{align*}
\mathtt{planf(} &\mathtt{pos\_hori(obj0,1)}, \mathtt{pos\_hori(obj0,3),}\mathtt{[move\_right,move\_right]).}
\end{align*}
This shows that
NEMESYS is able to perceive objects from an image, reason about the image, and edit the image through planning. To the best of our knowledge, this is the first differentiable neuro-symbolic system equipped with all of these abilities. We provide more Visual Concept Repairing tasks in Appendix~\ref{plan}.

\subsection{Differentiable Causal Reasoning}
As the last meta interpreter, we show that NEMESYS exhibits superior performance compared to the existing forward reasoning system by having the causal reasoning ability. 
Notably, given a causal Bayesian network, NEMESYS can perform the $\mathtt{do}$ operation (deleting the incoming edges of a node)~\citep{pearl2012docalculus} on arbitrary nodes and perform causal reasoning without the necessity of re-executing the entire system, which is made possible through meta-level programming.

\begin{wrapfigure}{t}{0.6\textwidth}
     \centering
     \includegraphics[trim=0 310 190 98, clip,width=0.6\textwidth]{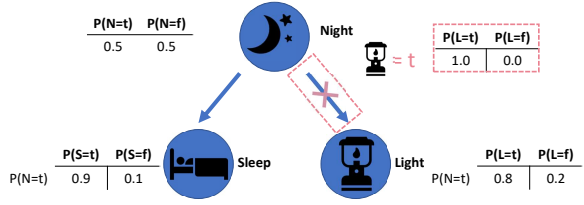}
          \caption{\textbf{Performing differentiable causal reasoning and learning using NEMESYS.} Given a causal Bayesian network, NEMESYS can easily perform the do operation (delete incoming edges) on arbitrary nodes and capture the causal effects on different nodes (for example, the probability of the node $\mathtt{Light}$ after intervening) without rerunning the entire system. Furthermore, NEMESYS is able to learn the unobserved $\mathtt{do}$ operation with its corresponding value using gradient descent based on the given causal graph and observed data. (Best viewed in color)}
     \label{fig:causal}
\end{wrapfigure}




The $\mathtt{do}$ operator, denoted as $\mathtt{do(X)}$, is used to represent an intervention on a particular variable $\mathtt{X}$ in a causal learning system, regardless of the actual value of the variable. 
For example, Fig.~\ref{fig:causal} shows a causal Bayesian network with three nodes and the probability distribution of the nodes before and after the $\mathtt{do}$ operation. 
To investigate how the node $\mathtt{Light}$ affects the rest of the system, we firstly cut the causal relationship between the node $\mathtt{Light}$ and all its parent nodes, then we assign a new value to the node and we investigate the probability of other nodes. To enable NEMESYS to perform a $\mathtt{do}$ operation on the node $\mathtt{Light}$, we begin by representing the provided causal Bayesian network in Fig.~\ref{fig:causal} using:
\begin{align*}
\mathtt{0.5}\texttt{:}\ \mathtt{Night}.\quad \mathtt{0.9}\texttt{:}\ \mathtt{Sleep}\texttt{:-}\mathtt{Night}.\quad \mathtt{0.8}\texttt{:}\ \mathtt{Light}\texttt{:-}\mathtt{Night}.
\end{align*}
where the number of an atom indicates the probability of the atom being true, and the number of a clause indicates the conditional probability of the head being true given the body being true.

We reuse the meta predicate $\mathtt{assert\_probs/1/[atom]}$ and introduce three new meta predicates: $\mathtt{prob/1/[atom]}$, $\mathtt{probs/1/[atoms]}$ and $\mathtt{probs\_do/1/[atoms,atom]}$. Since we cannot deal with the uncountable, infinite real numbers within our logic, we make use of the weight associated with ground meta atoms to represent the probability of the atom. 
For example, we use the weight of the meta atom $\mathtt{prob(Atom)}$ to represent the probability of the atom $\mathtt{Atom}$. We use the weight of the meta atom $\mathtt{probs(Atoms)}$ to represent the joint probability of a list of atoms $\mathtt{Atoms}$, and the weight of $\mathtt{probs\_do(AtomA,AtomB)}$ to represent the probability of the atom $\mathtt{AtomA}$ after performing the do operation $\mathtt{do(AtomB)}$. We modify the meta interpreter as:
\begin{align*}
&\mathtt{prob(Head)}\texttt{:-}\mathtt{assert\_probs((Head\texttt{:-}Body))},\mathtt{probs(Body).} \\
&\mathtt{probs([Body\_Atom,Body])}\texttt{:-}\mathtt{prob(Body\_Atom)},\mathtt{probs(Body)}. \\
&\mathtt{prob(Body\_Atom)}\texttt{:-}\mathtt{assert\_probs(Body\_Atom)}.  \\
&\mathtt{probs\_do(Atom,Atom)}\texttt{:-}\mathtt{do(Atom)}. \\
&\mathtt{probs\_do(Head,Atom)}\texttt{:-}\mathtt{assert\_probs((Head\texttt{:-}Body))},\mathtt{probs\_do(Body,Atom)}.\\
&\mathtt{probs\_do([Head,Tail],Atom)}\texttt{:-}\mathtt{probs\_do(Head,Atom)},\mathtt{probs\_do(Tail,Atom)}.\\
&\mathtt{probs\_do([Head,Tail],Head)}\texttt{:-}\mathtt{probs\_do(Head,Head)},\mathtt{probs(Tail)}.\\
&\mathtt{probs\_do(Atom,Do\_atom)}\texttt{:-}\mathtt{prob(Atom)}.
\end{align*}
where the first three rules calculate the probability of a node before the intervention, the joint probability is approximated using the first and second rule by iteratively multiplying each atom. The fourth rule assigns the probability of the atom $\mathtt{Atom}$ using the $\mathtt{do}$ operation. The fifth to the eighth calculate the probability after the $\mathtt{do}$ intervention by looping over each atom and multiplying them. 

For example, after performing $\mathtt{do(Light)}$ and setting the probability of $\mathtt{Light}$ as $1.0$. NEMESYS returns 
the weight of  $\mathtt{probs\_do(Light,Light)}$ as the probability of the node $\mathtt{Light}$ (Fig.~\ref{fig:causal} red box) after the intervention $\mathtt{do(Light)}$. 



\subsection{Gradient-based Learning in NEMESYS}

NEMESYS alleviates the limitations of frameworks such as DeepProbLog~\citep{manhaeve2018deepproblog} by having the ability of not only performing differentiable parameter learning but also supporting differentiable structure learning (in our experiment, NEMESYS learns the weights of the meta rules while adapting to solve different tasks). We now introduce the learning ability of NEMESYS. 


\subsubsection{Parameter Learning}
Consider a scenario in which a patient can only experience effective treatment when two types of medicine synergize, with the effectiveness contingent on the dosage of each drug. Suppose we have known the dosages of two medicines and the causal impact of the medicines on the patient, however, the observed effectiveness does not align with expectations. It is certain that some interventions have occurred in the medicine-patient causal structure (such as an incorrect dosage of one medicine, which will be treated as an intervention using the $\mathtt{do}$ operation). However, the specific node (patient or the medicines) on which the $\mathtt{do}$ operation is executed, and the values assigned to the $\mathtt{do}$ operator remain unknown. Conducting additional experiments on patients by altering medicine dosages to uncover the $\mathtt{do}$ operation is both unethical and dangerous.

With NEMESYS at hand, we can easily learn the unobserved $\mathtt{do}$ operation with its assigned value. We abstract the problem using a three-node causal Bayesian network: $$\mathtt{1.0: medicine\_a.} \quad \mathtt{1.0: medicine\_b.} \quad
\mathtt{0.9: patient} \texttt{:-} \mathtt{medicine\_a, medicine\_b.}$$
where the number of the atoms indicates the dosage of each medicine, and the number of the clause indicates the conditional probability of the effectiveness of the patient given these two medicines.  Suppose there is only one unobserved $\mathtt{do}$ operation.

\begin{figure}[t]
    \centering
    \includegraphics[width=0.495\textwidth]{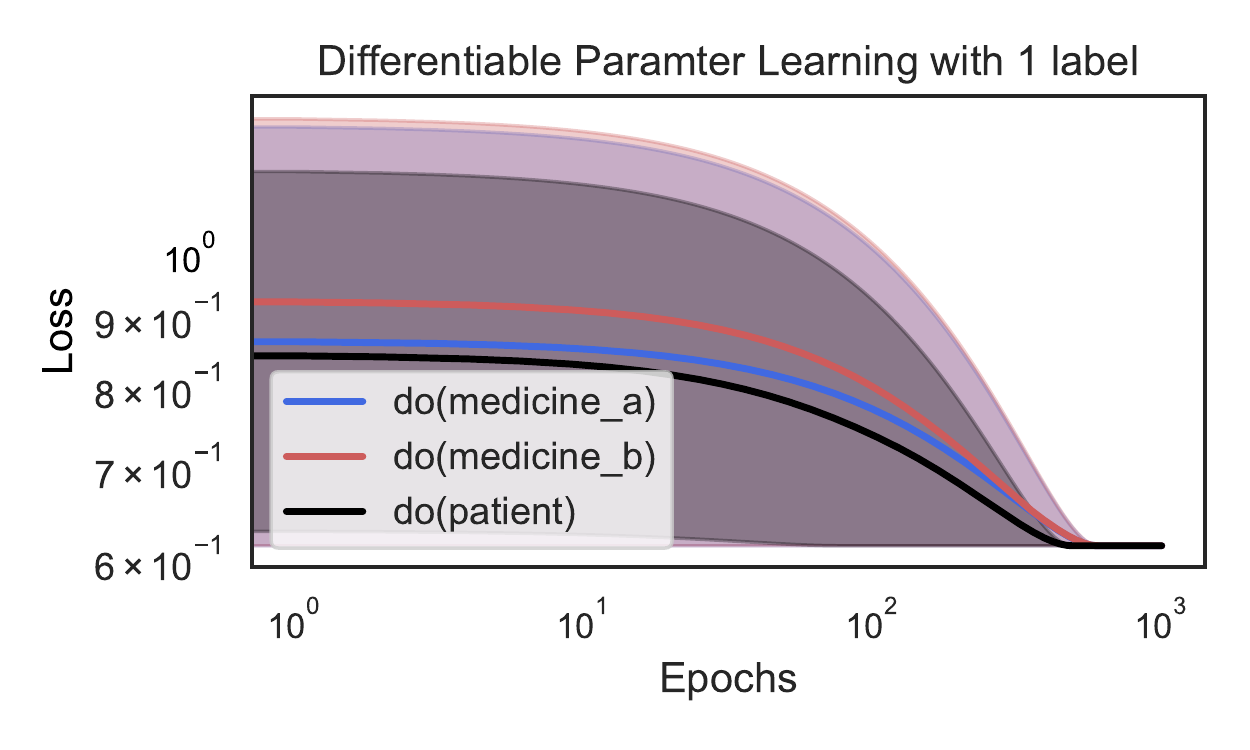}
    \includegraphics[width=0.495\textwidth]{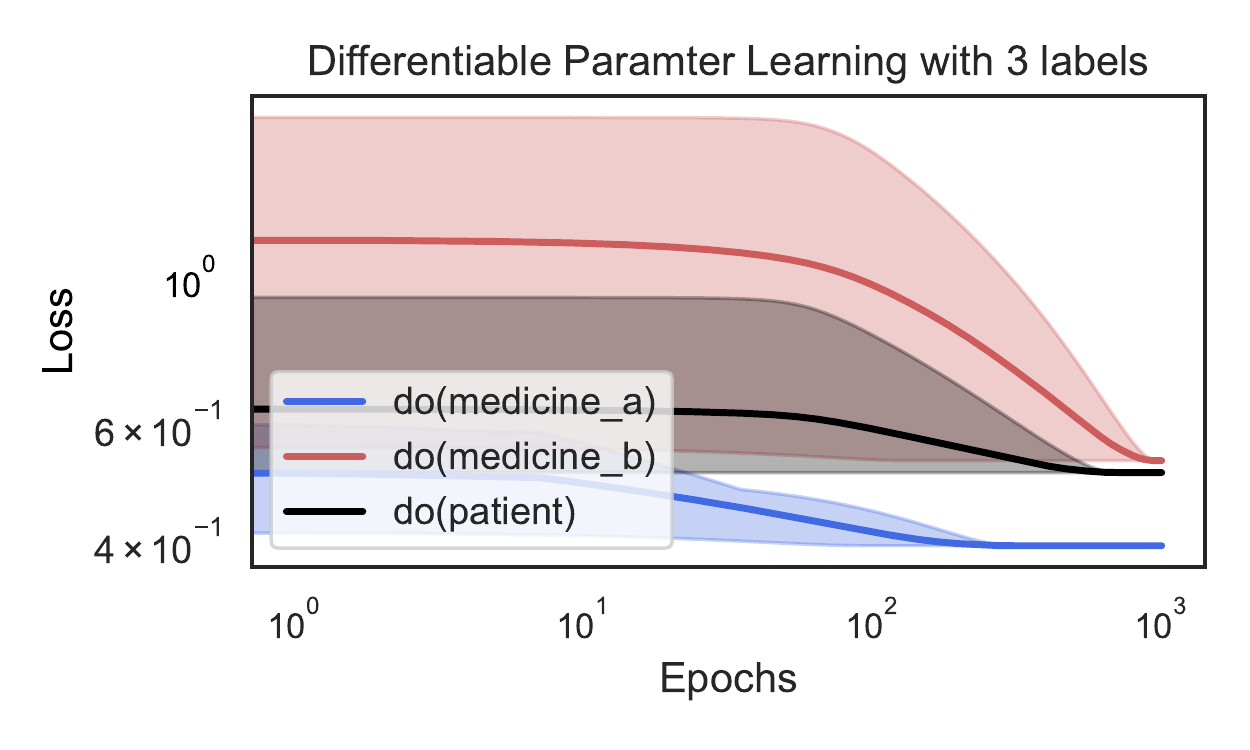}

    \caption{\textbf{NEMESYS performs differentiable parameter learning using gradient descent.} Based on the given data (one or three targets), NEMESYS is asked to learn the correct $\mathtt{do}$ operator and its corresponding value (which we didn't show in the images). The loss curve is averaged on three runs. The shadow area indicates the min and max number of the three runs. (Best viewed in color) }
    \label{fig:losscurve}
\end{figure}

To learn the unknown $\mathtt{do}$ operation, we define the loss as the Binary Cross Entropy (BCE) loss between the observed probability $\mathbf{p}_{target}$ 
and the predicted probability of the target atom $\mathbf{p}_{predicted}$. The predicted probability 
$\mathbf{p}_{predicted}$ is computed as: $ \mathbf{p}_{predicted}=\mathbf{v}^{(T)}\left[I_{\mathcal{G}}(\operatorname{target\_atom})\right]$, where $I_{\mathcal{G}}(x)$ is a function that returns the index of target atom in $\mathcal{G}$, $\mathbf{v}[i]$ is the $i$-th element of $\mathbf{v}$. $\mathbf{v}^{(T)}$ is the valuation tensor computed by $T$-step forward reasoning based on the initial valuation tensor $\mathbf{v}^{(0)}$, which is composed of the initial valuation of $\mathtt{do}$ and other meta ground atoms. 
Since the valuation of $\mathtt{do}$ atom is the only changing parameter, we set the gradient of other parameters as $0$. We minimize the loss w.r.t. $\mathtt{do(X)}$:
$
        \underset{\mathtt{do(X)}}{\mathtt{minimize}} \quad  \mathtt{L_{loss}}= \mathtt{BCE}(\mathbf{p}_{target},\mathbf{p}_{predicted}\mathtt{(do(X)))}.
$
Fig.~\ref{fig:losscurve} summarizes the loss curve of the three $\mathtt{do}$ operators during learning using one target (Fig.~\ref{fig:losscurve} left) and three targets (Fig.~\ref{fig:losscurve} right). For the three targets experiment, $\mathbf{p}_{target}$
consists of three observed probabilities (the effectiveness of the patient and the dosages of two medicines), for the experiment with one target, $\mathbf{p}_{target}$ only consists of the observed the effectiveness of the patient.

We randomly initialize the probability of the three $\mathtt{do}$ operators and choose the one, which achieves the lowest loss as the right $\mathtt{do}$ operator. In the three targets experiment, the blue curve achieves the lowest loss, with its corresponding value converges to the ground truth value, while in the one target experiment, three $\mathtt{do}$ operators achieve equivalent performance. We provide the value curves of three $\mathtt{do}$ operators and the ground truth $\mathtt{do}$ operator with its value in Appendix~\ref{diff_do_value_curve}. 


\begin{figure}[t]
    \centering
    \includegraphics[trim=8 82 5 80, clip, width=\textwidth]{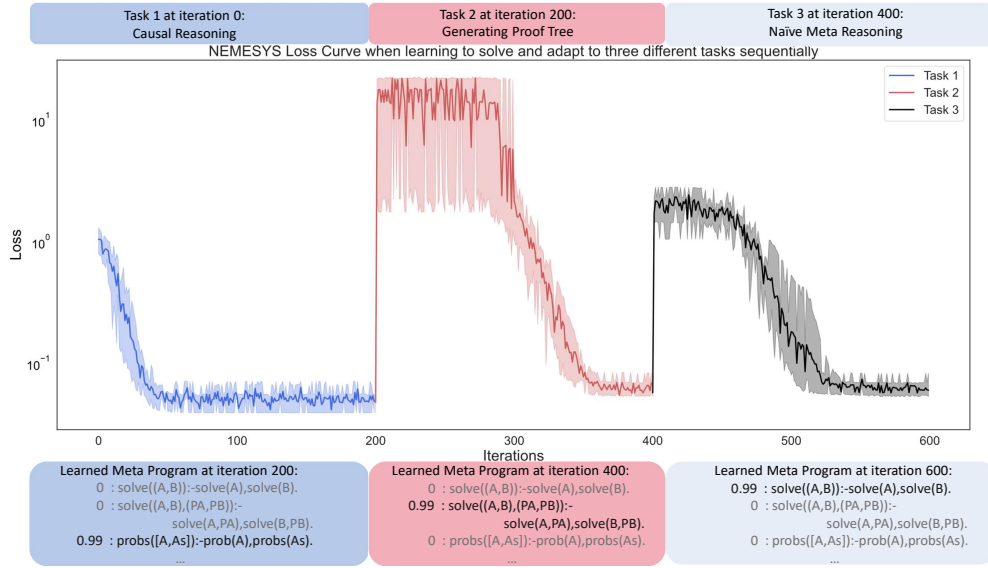}
    \caption{\textbf{NEMESYS can learn to solve and adapt itself to different tasks during learning using gradient descent.} In this experiment, we train NEMESYS to solve three different tasks: causal reasoning, generating proof trees and naive meta reasoning sequentially (each task is represented by a unique color encoding). The loss curve is averaged on five runs, with the shadow area indicating the minimum and maximum number of the five runs. For readability, the learned complete meta program is not shown in the image. (Best viewed in color)}
    \label{fig:multitask}
\end{figure}

\subsubsection{Structure Learning}

Besides parameter learning, NEMESYS can also perform differentiable structure learning (we provide the candidate meta rules and learn the weights of these meta rules using gradient descent). In this experiment, different tasks are presented at distinct time steps throughout the learning process. NEMESYS is tasked with acquiring the ability to solve and adapt to these diverse tasks.


Following Sec.~\ref{section:differentiability}, we make use of the meta rule weight matrix $\mathbf{W} = [ {\bf w}_1, \ldots, {\bf w}_M ]$ to select the rules. We take the \emph{softmax} of each weight vector ${\bf w}_j \in \mathbf{W}$ to choose $M$ meta rules out of $C$ meta rules. 
To adapt to different tasks, the weight matrix $\mathbf{W}$ is learned based on the loss, which is defined as the BCE loss between the probability of the target $\mathbf{p}_{target}$ and the predicted probability $\mathbf{p}_{predicted}$, where $\mathbf{p}_{predicted}$ is the probability of the target atoms calculated using the learned program. $\mathbf{p}_{predicted}$ is computed as: $ \mathbf{p}_{predicted}=\mathbf{v}^{(T)}\left[I_{\mathcal{G}}(\operatorname{target\_atoms})\right]$, where $I_{\mathcal{G}}(x)$ is a function that returns the indexes of target atoms in $\mathcal{G}$, $\mathbf{v}[i]$ is the $i$-th element of $\mathbf{v}$ and $\mathbf{v}^{(T)}$ is the valuation tensor computed by $T$-step forward reasoning. We minimize the loss w.r.t. the weight matrix $\mathbf{W}$: 
$
        \underset{\mathbf{W}}{\mathtt{minimize}} \quad  \mathtt{L_{loss}}= \mathtt{BCE}(\mathbf{p}_{target},\mathbf{p}_{predicted}(\mathbf{W})).
$

We randomly initialize the weight matrix $\mathbf{W}$, and update the weights using gradient descent. We set the target $\mathbf{p}_{target}$ using positive target atoms and negative target atoms. 
For example, suppose we have the naive meta reasoning and generating proof tree as two tasks. To learn a program to generate the proof tree, we use the proof tree meta rules to generate positive examples, and use the naive meta rules to generate the negative examples. 

We ask NEMESYS to solve three different tasks sequentially, which is 
initially, 
calculating probabilities using the first three rules of causal reasoning, then executing naive meta-reasoning. Finally, generating a proof tree.
We set the program size to three 
and randomly initialize the weight matrix. 
Fig.~\ref{fig:multitask} shows the learning process of NEMESYS which can automatically adapt to solve these three different tasks. 
 We provide the accuracy curve and the candidate rules with the learned weights in Appendix~\ref{multitask}. We also compare NEMESYS with the baseline method DeepProbLog~\citep{manhaeve2018deepproblog} (cf.~Table.~\ref{tab:multi_tasks}). Due to the limitations of DeepProbLog in adapting (meta) rule weights during learning, we initialize DeepProbLog with three variants as our baseline comparisons. The first variant involves fixed task 1 meta rule weights ($\mathtt{1.0}$), with task 2 and task 3 meta rule weights being randomly initialized. In the second variant, task 2 meta rule weights are fixed ($\mathtt{1.0}$), while task 1 and task 3 meta rule weights are randomly initialized, and this pattern continues for the subsequent variant. We provide NEMESYS with the same candidate meta rules, however, with randomly initialize  weights. We compute the accuracy at iteration $\mathtt{200}$, $\mathtt{400}$ and $\mathtt{600}$. 
\subsection{Discussion}\label{sec5}
While NEMESYS achieves impressive results, it is worth considering some of the limitations of this work.
In our current experiments for structure learning, candidates of meta-rules are provided. It is promising to integrate rule-learning techniques, \eg mode declarations, meta-interpretive learning, and a more sophisticated rule search, to learn from less prior. Another limitation lies in the calculation, since our system is not able to handle real number calculation, we make use of the weight associated with the atom to approximate the value and do the calculation.
\begin{table*}[t]
\small
\centering
\begin{tabular}{lccc}
              & Test Task1  & Test Task2 & Test Task3\\ 
DeepProbLog~(T1)        &100 $\bullet$    &14.29            &0            \\
DeepProbLog~(T2)         &0      &100 $\bullet$           &11.43            \\
DeepProbLog~(T3)           &68.57        &5.71          & 100 $\bullet$             \\
NEMESYS~(ours)  &100 $\bullet$   &100 $\bullet$        &100 $\bullet$          \\ 
\end{tabular}
\caption{\textbf{Performance (Accuracy; the higher, the better) on test split of three tasks.} We compare NEMESYS with baseline method DeepProbLog~\citep{manhaeve2018deepproblog} (with three variants). The accuracy is averaged on five runs. The best-performing models are denoted using $\bullet$.}
\label{tab:multi_tasks}
\end{table*}
\section{Conclusions}\label{sec6}

We proposed the framework of neuro-metasymbolic reasoning and learning.  We realized a differentiable meta interpreter using the differentiable implementation of first-order logic with meta predicates. This meta-interpreter, called NEMESYS, 
achieves various important functions on differentiable logic programming languages using meta-level programming.
We illustrated this on different tasks of visual reasoning, reasoning with explanations, reasoning with infinite loops, planning on visual scenes, performing the $\mathtt{do}$ operation within a causal Bayesian network and showed NEMESYS's gradient-based capability of parameter learning and structure learning.

NEMESYS provides several interesting avenues for future work.
One major limitation of NEMESYS is its scalability for large-scale meta programs. 
So far, we have mainly focused on specifying the syntax and semantics of new (domain-specific) differentiable logic programming languages, helping to ensure that the languages have some desired properties. 
In the future, one should also explore providing properties about programs written in a particular differentiable logic programming language and injecting the properties into deep neural networks via algorithmic supervision~\cite{petersen2021algorithmic_supervision}, as well as program synthesis.
Most importantly, since meta programs in NEMESYS are parameterized, and the reasoning mechanism is differentiable, one can realize differentiable meta-learning easily, \ie the reasoning system that learns how to perform reasoning better from experiences.

\vspace{1em}
\noindent
\textbf{Acknowledgements.}
This work was supported by the Hessian Ministry of Higher Education, Research, Science and the Arts (HMWK) cluster project “The Third Wave of AI”. The work has also benefited from the Hessian Ministry of Higher Education, Research, Science and the Arts (HMWK) cluster project “The Adaptive Mind” and the Federal Ministry for Economic Affairs and Climate Action (BMWK) AI lighthouse project “SPAICER” (01MK20015E), the EU ICT-48 Network of AI Research Excellence Center “TAILOR” (EU
Horizon 2020, GA No 952215), and the Collaboration Lab “AI in Construction” (AICO) with Nexplore/HochTief.

\bibliography{sn-bibliography}


\begin{thebibliography}{50}
\ifx \bisbn   \undefined \def \bisbn  #1{ISBN #1}\fi
\ifx \binits  \undefined \def \binits#1{#1}\fi
\ifx \bauthor  \undefined \def \bauthor#1{#1}\fi
\ifx \batitle  \undefined \def \batitle#1{#1}\fi
\ifx \bjtitle  \undefined \def \bjtitle#1{#1}\fi
\ifx \bvolume  \undefined \def \bvolume#1{\textbf{#1}}\fi
\ifx \byear  \undefined \def \byear#1{#1}\fi
\ifx \bissue  \undefined \def \bissue#1{#1}\fi
\ifx \bfpage  \undefined \def \bfpage#1{#1}\fi
\ifx \blpage  \undefined \def \blpage #1{#1}\fi
\ifx \burl  \undefined \def \burl#1{\textsf{#1}}\fi
\ifx \doiurl  \undefined \def \doiurl#1{\url{https://doi.org/#1}}\fi
\ifx \betal  \undefined \def \betal{\textit{et al.}}\fi
\ifx \binstitute  \undefined \def \binstitute#1{#1}\fi
\ifx \binstitutionaled  \undefined \def \binstitutionaled#1{#1}\fi
\ifx \bctitle  \undefined \def \bctitle#1{#1}\fi
\ifx \beditor  \undefined \def \beditor#1{#1}\fi
\ifx \bpublisher  \undefined \def \bpublisher#1{#1}\fi
\ifx \bbtitle  \undefined \def \bbtitle#1{#1}\fi
\ifx \bedition  \undefined \def \bedition#1{#1}\fi
\ifx \bseriesno  \undefined \def \bseriesno#1{#1}\fi
\ifx \blocation  \undefined \def \blocation#1{#1}\fi
\ifx \bsertitle  \undefined \def \bsertitle#1{#1}\fi
\ifx \bsnm \undefined \def \bsnm#1{#1}\fi
\ifx \bsuffix \undefined \def \bsuffix#1{#1}\fi
\ifx \bparticle \undefined \def \bparticle#1{#1}\fi
\ifx \barticle \undefined \def \barticle#1{#1}\fi
\bibcommenthead
\ifx \bconfdate \undefined \def \bconfdate #1{#1}\fi
\ifx \botherref \undefined \def \botherref #1{#1}\fi
\ifx \url \undefined \def \url#1{\textsf{#1}}\fi
\ifx \bchapter \undefined \def \bchapter#1{#1}\fi
\ifx \bbook \undefined \def \bbook#1{#1}\fi
\ifx \bcomment \undefined \def \bcomment#1{#1}\fi
\ifx \oauthor \undefined \def \oauthor#1{#1}\fi
\ifx \citeauthoryear \undefined \def \citeauthoryear#1{#1}\fi
\ifx \endbibitem  \undefined \def \endbibitem {}\fi
\ifx \bconflocation  \undefined \def \bconflocation#1{#1}\fi
\ifx \arxivurl  \undefined \def \arxivurl#1{\textsf{#1}}\fi
\csname PreBibitemsHook\endcsname

\bibitem[\protect\citeauthoryear{Ramesh et~al.}{2022}]{dalle2}
\begin{botherref}
\oauthor{\bsnm{Ramesh}, \binits{A.}},
\oauthor{\bsnm{Dhariwal}, \binits{P.}},
\oauthor{\bsnm{Nichol}, \binits{A.}},
\oauthor{\bsnm{Chu}, \binits{C.}},
\oauthor{\bsnm{Chen}, \binits{M.}}:
Hierarchical text-conditional image generation with clip latents.
arXiv Preprint:2204.0612
(2022)
\end{botherref}
\endbibitem

\bibitem[\protect\citeauthoryear{Stiennon et~al.}{2020}]{stiennon2020learning}
\begin{botherref}
\oauthor{\bsnm{Stiennon}, \binits{N.}},
\oauthor{\bsnm{Ouyang}, \binits{L.}},
\oauthor{\bsnm{Wu}, \binits{J.}},
\oauthor{\bsnm{Ziegler}, \binits{D.}},
\oauthor{\bsnm{Lowe}, \binits{R.}},
\oauthor{\bsnm{Voss}, \binits{C.}},
\oauthor{\bsnm{Radford}, \binits{A.}},
\oauthor{\bsnm{Amodei}, \binits{D.}},
\oauthor{\bsnm{Christiano}, \binits{P.F.}}:
Learning to summarize with human feedback.
Advances in Neural Information Processing Systems (NeurIPS)
(2020)
\end{botherref}
\endbibitem

\bibitem[\protect\citeauthoryear{Floridi and Chiriatti}{2020}]{floridi2020gpt}
\begin{barticle}
\bauthor{\bsnm{Floridi}, \binits{L.}},
\bauthor{\bsnm{Chiriatti}, \binits{M.}}:
\batitle{Gpt-3: Its nature, scope, limits, and consequences}.
\bjtitle{Minds and Machines}
\bvolume{30},
\bfpage{681}--\blpage{694}
(\byear{2020})
\end{barticle}
\endbibitem

\bibitem[\protect\citeauthoryear{Reed et~al.}{2022}]{reedgeneralist}
\begin{botherref}
\oauthor{\bsnm{Reed}, \binits{S.}},
\oauthor{\bsnm{Zolna}, \binits{K.}},
\oauthor{\bsnm{Parisotto}, \binits{E.}},
\oauthor{\bsnm{Colmenarejo}, \binits{S.G.}},
\oauthor{\bsnm{Novikov}, \binits{A.}},
\oauthor{\bsnm{Barth-maron}, \binits{G.}},
\oauthor{\bsnm{Gim{\'e}nez}, \binits{M.}},
\oauthor{\bsnm{Sulsky}, \binits{Y.}},
\oauthor{\bsnm{Kay}, \binits{J.}},
\oauthor{\bsnm{Springenberg}, \binits{J.T.}}, et al.:
A generalist agent.
Transactions on Machine Learning Research (TMLR)
(2022)
\end{botherref}
\endbibitem

\bibitem[\protect\citeauthoryear{Ackerman and
  Thompson}{2017}]{ackerman2017metareasoning}
\begin{barticle}
\bauthor{\bsnm{Ackerman}, \binits{R.}},
\bauthor{\bsnm{Thompson}, \binits{V.A.}}:
\batitle{Meta-reasoning: Monitoring and control of thinking and reasoning}.
\bjtitle{Trends in cognitive sciences}
\bvolume{21}(\bissue{8}),
\bfpage{607}--\blpage{617}
(\byear{2017})
\end{barticle}
\endbibitem

\bibitem[\protect\citeauthoryear{Costantini}{2002}]{Costantini2002meta}
\begin{bchapter}
\bauthor{\bsnm{Costantini}, \binits{S.}}:
\bctitle{Meta-reasoning: A survey}.
In: \bbtitle{Computational Logic: Logic Programming and Beyond}
(\byear{2002})
\end{bchapter}
\endbibitem

\bibitem[\protect\citeauthoryear{Griffiths et~al.}{2019}]{Griffiths2019meta}
\begin{barticle}
\bauthor{\bsnm{Griffiths}, \binits{T.L.}},
\bauthor{\bsnm{Callaway}, \binits{F.}},
\bauthor{\bsnm{Chang}, \binits{M.B.}},
\bauthor{\bsnm{Grant}, \binits{E.}},
\bauthor{\bsnm{Krueger}, \binits{P.M.}},
\bauthor{\bsnm{Lieder}, \binits{F.}}:
\batitle{Doing more with less: Meta-reasoning and meta-learning in humans and
  machines}.
\bjtitle{Current Opinion in Behavioral Sciences}
\bvolume{29},
\bfpage{24}--\blpage{30}
(\byear{2019})
\end{barticle}
\endbibitem

\bibitem[\protect\citeauthoryear{Russell and
  Wefald}{1991}]{russell1991metaprinciples}
\begin{barticle}
\bauthor{\bsnm{Russell}, \binits{S.}},
\bauthor{\bsnm{Wefald}, \binits{E.}}:
\batitle{Principles of metareasoning}.
\bjtitle{Artificial intelligence}
\bvolume{49}(\bissue{1-3}),
\bfpage{361}--\blpage{395}
(\byear{1991})
\end{barticle}
\endbibitem

\bibitem[\protect\citeauthoryear{Schmidhuber}{1987}]{schmidhuber1987evolutionary}
\begin{botherref}
\oauthor{\bsnm{Schmidhuber}, \binits{J.}}:
Evolutionary principles in self-referential learning, or on learning how to
  learn: the meta-meta-... hook.
PhD thesis,
Technische Universit{\"a}t M{\"u}nchen
(1987)
\end{botherref}
\endbibitem

\bibitem[\protect\citeauthoryear{Thrun and
  Pratt}{1998}]{Thrun1998learningtolearn}
\begin{bbook}
\bauthor{\bsnm{Thrun}, \binits{S.}},
\bauthor{\bsnm{Pratt}, \binits{L.}}:
\bbtitle{Learning to Learn: Introduction and Overview},
pp. \bfpage{3}--\blpage{17}.
\bpublisher{Springer},
\blocation{Boston, MA}
(\byear{1998})
\end{bbook}
\endbibitem

\bibitem[\protect\citeauthoryear{Finn
  et~al.}{2017}]{Finn17modelagnostic_metalearning}
\begin{bchapter}
\bauthor{\bsnm{Finn}, \binits{C.}},
\bauthor{\bsnm{Abbeel}, \binits{P.}},
\bauthor{\bsnm{Levine}, \binits{S.}}:
\bctitle{Model-agnostic meta-learning for fast adaptation of deep networks}.
In: \bbtitle{Proceedings of the 34th International Conference on Machine
  Learning (ICML)}
(\byear{2017})
\end{bchapter}
\endbibitem

\bibitem[\protect\citeauthoryear{Hospedales
  et~al.}{2022}]{Hospedales22metalearning}
\begin{barticle}
\bauthor{\bsnm{Hospedales}, \binits{T.M.}},
\bauthor{\bsnm{Antoniou}, \binits{A.}},
\bauthor{\bsnm{Micaelli}, \binits{P.}},
\bauthor{\bsnm{Storkey}, \binits{A.J.}}:
\batitle{Meta-learning in neural networks: {A} survey}.
\bjtitle{{IEEE} Trans. Pattern Anal. Mach. Intell.}
\bvolume{44}(\bissue{9}),
\bfpage{5149}--\blpage{5169}
(\byear{2022})
\end{barticle}
\endbibitem

\bibitem[\protect\citeauthoryear{Kim et~al.}{2018}]{kim2018notsoclevr}
\begin{botherref}
\oauthor{\bsnm{Kim}, \binits{J.}},
\oauthor{\bsnm{Ricci}, \binits{M.}},
\oauthor{\bsnm{Serre}, \binits{T.}}:
Not-so-clevr: learning same--different relations strains feedforward neural
  networks.
Interface focus
(2018)
\end{botherref}
\endbibitem

\bibitem[\protect\citeauthoryear{Stammer et~al.}{2021}]{Stammer21clevrhans}
\begin{bchapter}
\bauthor{\bsnm{Stammer}, \binits{W.}},
\bauthor{\bsnm{Schramowski}, \binits{P.}},
\bauthor{\bsnm{Kersting}, \binits{K.}}:
\bctitle{Right for the right concept: Revising neuro-symbolic concepts by
  interacting with their explanations}.
In: \bbtitle{Proceedings of the {IEEE} Conference on Computer Vision and
  Pattern Recognition ({CVPR})}
(\byear{2021})
\end{bchapter}
\endbibitem

\bibitem[\protect\citeauthoryear{Shindo et~al.}{2021}]{Shindo21nsfr}
\begin{botherref}
\oauthor{\bsnm{Shindo}, \binits{H.}},
\oauthor{\bsnm{Dhami}, \binits{D.S.}},
\oauthor{\bsnm{Kersting}, \binits{K.}}:
Neuro-symbolic forward reasoning.
arXiv Preprint:2110.09383
(2021)
\end{botherref}
\endbibitem

\bibitem[\protect\citeauthoryear{Evans and Grefenstette}{2018}]{Evans18}
\begin{barticle}
\bauthor{\bsnm{Evans}, \binits{R.}},
\bauthor{\bsnm{Grefenstette}, \binits{E.}}:
\batitle{Learning explanatory rules from noisy data}.
\bjtitle{J. Artif. Intell. Res.}
\bvolume{61},
\bfpage{1}--\blpage{64}
(\byear{2018})
\end{barticle}
\endbibitem

\bibitem[\protect\citeauthoryear{Shindo et~al.}{2021}]{Shindo21aaai}
\begin{bchapter}
\bauthor{\bsnm{Shindo}, \binits{H.}},
\bauthor{\bsnm{Nishino}, \binits{M.}},
\bauthor{\bsnm{Yamamoto}, \binits{A.}}:
\bctitle{Differentiable inductive logic programming for structured examples}.
In: \bbtitle{Proceedings of the 35th {AAAI} Conference on Artificial
  Intelligence (AAAI)}
(\byear{2021})
\end{bchapter}
\endbibitem

\bibitem[\protect\citeauthoryear{Johnson et~al.}{2017}]{Johnson17clevr}
\begin{botherref}
\oauthor{\bsnm{Johnson}, \binits{J.}},
\oauthor{\bsnm{Hariharan}, \binits{B.}},
\oauthor{\bsnm{Maaten}, \binits{L.}},
\oauthor{\bsnm{Fei-Fei}, \binits{L.}},
\oauthor{\bsnm{Zitnick}, \binits{C.L.}},
\oauthor{\bsnm{Girshick}, \binits{R.B.}}:
Clevr: A diagnostic dataset for compositional language and elementary visual
  reasoning.
Proceedings of the IEEE Conference on Computer Vision and Pattern Recognition
  (CVPR)
(2017)
\end{botherref}
\endbibitem

\bibitem[\protect\citeauthoryear{Holzinger et~al.}{2019}]{Holzinger19IQ}
\begin{bchapter}
\bauthor{\bsnm{Holzinger}, \binits{A.}},
\bauthor{\bsnm{Kickmeier-Rust}, \binits{M.}},
\bauthor{\bsnm{M\"{u}ller}, \binits{H.}}:
\bctitle{Kandinsky patterns as iq-test for machine learning}.
In: \bbtitle{Proceedings of the 3rd International Cross-Domain Conference for
  Machine Learning and Knowledge Extraction (CD-MAKE)}
(\byear{2019})
\end{bchapter}
\endbibitem

\bibitem[\protect\citeauthoryear{Müller and Holzinger}{2021}]{Mueller21}
\begin{barticle}
\bauthor{\bsnm{Müller}, \binits{H.}},
\bauthor{\bsnm{Holzinger}, \binits{A.}}:
\batitle{Kandinsky patterns}.
\bjtitle{Artificial Intelligence}
\bvolume{300},
\bfpage{103546}
(\byear{2021})
\end{barticle}
\endbibitem

\bibitem[\protect\citeauthoryear{Manhaeve
  et~al.}{2018}]{manhaeve2018deepproblog}
\begin{botherref}
\oauthor{\bsnm{Manhaeve}, \binits{R.}},
\oauthor{\bsnm{Dumancic}, \binits{S.}},
\oauthor{\bsnm{Kimmig}, \binits{A.}},
\oauthor{\bsnm{Demeester}, \binits{T.}},
\oauthor{\bsnm{De~Raedt}, \binits{L.}}:
Deepproblog: Neural probabilistic logic programming.
Advances in Neural Information Processing Systems (NeurIPS)
(2018)
\end{botherref}
\endbibitem

\bibitem[\protect\citeauthoryear{Rockt{\"a}schel and
  Riedel}{2017}]{rocktaschel2017end}
\begin{botherref}
\oauthor{\bsnm{Rockt{\"a}schel}, \binits{T.}},
\oauthor{\bsnm{Riedel}, \binits{S.}}:
End-to-end differentiable proving.
Advances in neural information processing systems
\textbf{30}
(2017)
\end{botherref}
\endbibitem

\bibitem[\protect\citeauthoryear{Cunnington et~al.}{2023}]{cunnington2023ffnsl}
\begin{barticle}
\bauthor{\bsnm{Cunnington}, \binits{D.}},
\bauthor{\bsnm{Law}, \binits{M.}},
\bauthor{\bsnm{Lobo}, \binits{J.}},
\bauthor{\bsnm{Russo}, \binits{A.}}:
\batitle{Ffnsl: feed-forward neural-symbolic learner}.
\bjtitle{Machine Learning}
\bvolume{112}(\bissue{2}),
\bfpage{515}--\blpage{569}
(\byear{2023})
\end{barticle}
\endbibitem

\bibitem[\protect\citeauthoryear{Shindo et~al.}{2023}]{shindo2023alphailp}
\begin{barticle}
\bauthor{\bsnm{Shindo}, \binits{H.}},
\bauthor{\bsnm{Pfanschilling}, \binits{V.}},
\bauthor{\bsnm{Dhami}, \binits{D.S.}},
\bauthor{\bsnm{Kersting}, \binits{K.}}:
\batitle{$\alpha$ilp: thinking visual scenes as differentiable logic programs}.
\bjtitle{Machine Learning}
\bvolume{112},
\bfpage{1465}--\blpage{1497}
(\byear{2023})
\end{barticle}
\endbibitem

\bibitem[\protect\citeauthoryear{Huang et~al.}{2021}]{huang2021scallop}
\begin{botherref}
\oauthor{\bsnm{Huang}, \binits{J.}},
\oauthor{\bsnm{Li}, \binits{Z.}},
\oauthor{\bsnm{Chen}, \binits{B.}},
\oauthor{\bsnm{Samel}, \binits{K.}},
\oauthor{\bsnm{Naik}, \binits{M.}},
\oauthor{\bsnm{Song}, \binits{L.}},
\oauthor{\bsnm{Si}, \binits{X.}}:
Scallop: From probabilistic deductive databases to scalable differentiable
  reasoning.
Advances in Neural Information Processing Systems (NeurIPS)
(2021)
\end{botherref}
\endbibitem

\bibitem[\protect\citeauthoryear{Yang et~al.}{2020}]{Yang2020neuroasp}
\begin{bchapter}
\bauthor{\bsnm{Yang}, \binits{Z.}},
\bauthor{\bsnm{Ishay}, \binits{A.}},
\bauthor{\bsnm{Lee}, \binits{J.}}:
\bctitle{Neurasp: Embracing neural networks into answer set programming}.
In: \bbtitle{Proceedings of the 29th International Joint Conference on
  Artificial Intelligence, (IJCAI)}
(\byear{2020})
\end{bchapter}
\endbibitem

\bibitem[\protect\citeauthoryear{Pearl}{2009}]{pearl2009causality}
\begin{bbook}
\bauthor{\bsnm{Pearl}, \binits{J.}}:
\bbtitle{Causality}.
\bpublisher{Cambridge university press},
\blocation{Cambridge}
(\byear{2009})
\end{bbook}
\endbibitem

\bibitem[\protect\citeauthoryear{Pearl}{2012}]{pearl2012docalculus}
\begin{bchapter}
\bauthor{\bsnm{Pearl}, \binits{J.}}:
\bctitle{The do-calculus revisited}.
In: \bbtitle{Proceedings of the 28th Conference on Uncertainty in Artificial
  Intelligence (UAI)}
(\byear{2012})
\end{bchapter}
\endbibitem

\bibitem[\protect\citeauthoryear{Russell and Norvig}{2009}]{Russel09}
\begin{bbook}
\bauthor{\bsnm{Russell}, \binits{S.}},
\bauthor{\bsnm{Norvig}, \binits{P.}}:
\bbtitle{Artificial Intelligence: A Modern Approach},
\bedition{3rd} edn.
\bpublisher{Prentice Hall Press},
\blocation{Hoboken, New Jersey}
(\byear{2009})
\end{bbook}
\endbibitem

\bibitem[\protect\citeauthoryear{Jiang and Luo}{2019}]{Jiang19logicrl}
\begin{bchapter}
\bauthor{\bsnm{Jiang}, \binits{Z.}},
\bauthor{\bsnm{Luo}, \binits{S.}}:
\bctitle{Neural logic reinforcement learning}.
In: \bbtitle{Proceedings of the 36th International Conference on Machine
  Learning (ICML)}
(\byear{2019})
\end{bchapter}
\endbibitem

\bibitem[\protect\citeauthoryear{Delfosse
  et~al.}{2023}]{delfosse2023interpretable}
\begin{botherref}
\oauthor{\bsnm{Delfosse}, \binits{Q.}},
\oauthor{\bsnm{Shindo}, \binits{H.}},
\oauthor{\bsnm{Dhami}, \binits{D.}},
\oauthor{\bsnm{Kersting}, \binits{K.}}:
Interpretable and explainable logical policies via neurally guided symbolic
  abstraction.
arXiv preprint arXiv:2306.01439
(2023)
\end{botherref}
\endbibitem

\bibitem[\protect\citeauthoryear{Maes and Nardi}{1988}]{Maes1998Introspection}
\begin{bbook}
\bauthor{\bsnm{Maes}, \binits{P.}},
\bauthor{\bsnm{Nardi}, \binits{D.}}:
\bbtitle{Meta-Level Architectures and Reflection}.
\bpublisher{Elsevier Science Inc.},
\blocation{USA}
(\byear{1988})
\end{bbook}
\endbibitem

\bibitem[\protect\citeauthoryear{Lloyd}{1984}]{Lloyd84foundation}
\begin{bbook}
\bauthor{\bsnm{Lloyd}, \binits{J.W.}}:
\bbtitle{Foundations of Logic Programming, 1st Edition}.
\bpublisher{Springer},
\blocation{Heidelberg}
(\byear{1984})
\end{bbook}
\endbibitem

\bibitem[\protect\citeauthoryear{Hill and Gallagher}{1998}]{Hill1998metalogic}
\begin{bbook}
\bauthor{\bsnm{Hill}, \binits{P.M.}},
\bauthor{\bsnm{Gallagher}, \binits{J.}}:
\bbtitle{{Meta-Programming in Logic Programming}}.
\bpublisher{Oxford University Press},
\blocation{Oxford}
(\byear{1998})
\end{bbook}
\endbibitem

\bibitem[\protect\citeauthoryear{Pettorossi}{1992}]{Pettorossi1992metaproglogic}
\begin{bbook}
\beditor{\bsnm{Pettorossi}, \binits{A.}} (ed.):
\bbtitle{Proceedings of the 3rd International Workshop of Meta-Programming in
  Logic, (META)}.
\bsertitle{Lecture Notes in Computer Science},
vol. \bseriesno{649}
(\byear{1992})
\end{bbook}
\endbibitem

\bibitem[\protect\citeauthoryear{Apt and Turini}{1995}]{Apt1995metalogic}
\begin{bbook}
\bauthor{\bsnm{Apt}, \binits{K.R.}},
\bauthor{\bsnm{Turini}, \binits{F.}}:
\bbtitle{Meta-Logics and Logic Programming}.
\bpublisher{MIT Press (MA)},
\blocation{Massachusett}
(\byear{1995})
\end{bbook}
\endbibitem

\bibitem[\protect\citeauthoryear{Sterling and
  Shapiro}{1994}]{sterling1994artofprolog}
\begin{bbook}
\bauthor{\bsnm{Sterling}, \binits{L.}},
\bauthor{\bsnm{Shapiro}, \binits{E.Y.}}:
\bbtitle{The Art of Prolog: Advanced Programming Techniques}.
\bpublisher{MIT press},
\blocation{Massachusett}
(\byear{1994})
\end{bbook}
\endbibitem

\bibitem[\protect\citeauthoryear{Muggleton et~al.}{2014a}]{muggleton2014meta}
\begin{barticle}
\bauthor{\bsnm{Muggleton}, \binits{S.H.}},
\bauthor{\bsnm{Lin}, \binits{D.}},
\bauthor{\bsnm{Pahlavi}, \binits{N.}},
\bauthor{\bsnm{Tamaddoni-Nezhad}, \binits{A.}}:
\batitle{Meta-interpretive learning: application to grammatical inference}.
\bjtitle{Machine learning}
\bvolume{94},
\bfpage{25}--\blpage{49}
(\byear{2014})
\end{barticle}
\endbibitem

\bibitem[\protect\citeauthoryear{Muggleton
  et~al.}{2014b}]{muggleton2014metabayes}
\begin{botherref}
\oauthor{\bsnm{Muggleton}, \binits{S.H.}},
\oauthor{\bsnm{Lin}, \binits{D.}},
\oauthor{\bsnm{Chen}, \binits{J.}},
\oauthor{\bsnm{Tamaddoni-Nezhad}, \binits{A.}}:
Metabayes: Bayesian meta-interpretative learning using higher-order stochastic
  refinement.
Proceedings of the 24th International Conference on Inductive Logic Programming
  (ILP)
(2014)
\end{botherref}
\endbibitem

\bibitem[\protect\citeauthoryear{Muggleton et~al.}{2015}]{muggleton2015meta}
\begin{barticle}
\bauthor{\bsnm{Muggleton}, \binits{S.H.}},
\bauthor{\bsnm{Lin}, \binits{D.}},
\bauthor{\bsnm{Tamaddoni-Nezhad}, \binits{A.}}:
\batitle{Meta-interpretive learning of higher-order dyadic datalog: Predicate
  invention revisited}.
\bjtitle{Machine Learning}
\bvolume{100},
\bfpage{49}--\blpage{73}
(\byear{2015})
\end{barticle}
\endbibitem

\bibitem[\protect\citeauthoryear{Cuturi and Blondel}{2017}]{cuturi2017soft}
\begin{bchapter}
\bauthor{\bsnm{Cuturi}, \binits{M.}},
\bauthor{\bsnm{Blondel}, \binits{M.}}:
\bctitle{Soft-dtw: a differentiable loss function for time-series}.
In: \bbtitle{Proceedings of the 34th International Conference on Machine
  Learning (ICML)}
(\byear{2017})
\end{bchapter}
\endbibitem

\bibitem[\protect\citeauthoryear{Redmon et~al.}{2016}]{YOLO}
\begin{bchapter}
\bauthor{\bsnm{Redmon}, \binits{J.}},
\bauthor{\bsnm{Divvala}, \binits{S.}},
\bauthor{\bsnm{Girshick}, \binits{R.}},
\bauthor{\bsnm{Farhadi}, \binits{A.}}:
\bctitle{You only look once: Unified, real-time object detection}.
In: \bbtitle{Proceedings of the IEEE Conference on Computer Vision and Pattern
  Recognition (CVPR)}
(\byear{2016})
\end{bchapter}
\endbibitem

\bibitem[\protect\citeauthoryear{Holzinger et~al.}{2021}]{Holzinger21}
\begin{botherref}
\oauthor{\bsnm{Holzinger}, \binits{A.}},
\oauthor{\bsnm{Saranti}, \binits{A.}},
\oauthor{\bsnm{M{\"{u}}ller}, \binits{H.}}:
Kandinsky patterns - an experimental exploration environment for pattern
  analysis and machine intelligence.
arXiv Preprint:2103.00519
(2021)
\end{botherref}
\endbibitem

\bibitem[\protect\citeauthoryear{He et~al.}{2016}]{He_2016_Residualnetwork}
\begin{bchapter}
\bauthor{\bsnm{He}, \binits{K.}},
\bauthor{\bsnm{Zhang}, \binits{X.}},
\bauthor{\bsnm{Ren}, \binits{S.}},
\bauthor{\bsnm{Sun}, \binits{J.}}:
\bctitle{Deep residual learning for image recognition}.
In: \bbtitle{Proceedings of the IEEE Conference on Computer Vision and Pattern
  Recognition (CVPR)}
(\byear{2016})
\end{bchapter}
\endbibitem

\bibitem[\protect\citeauthoryear{Locatello et~al.}{2020}]{slotattention}
\begin{botherref}
\oauthor{\bsnm{Locatello}, \binits{F.}},
\oauthor{\bsnm{Weissenborn}, \binits{D.}},
\oauthor{\bsnm{Unterthiner}, \binits{T.}},
\oauthor{\bsnm{Mahendran}, \binits{A.}},
\oauthor{\bsnm{Heigold}, \binits{G.}},
\oauthor{\bsnm{Uszkoreit}, \binits{J.}},
\oauthor{\bsnm{Dosovitskiy}, \binits{A.}},
\oauthor{\bsnm{Kipf}, \binits{T.}}:
Object-centric learning with slot attention.
Advances in Neural Information Processing Systems (NeurIPS)
(2020)
\end{botherref}
\endbibitem

\bibitem[\protect\citeauthoryear{Lee et~al.}{2019}]{Lee19settransformer}
\begin{bchapter}
\bauthor{\bsnm{Lee}, \binits{J.}},
\bauthor{\bsnm{Lee}, \binits{Y.}},
\bauthor{\bsnm{Kim}, \binits{J.}},
\bauthor{\bsnm{Kosiorek}, \binits{A.}},
\bauthor{\bsnm{Choi}, \binits{S.}},
\bauthor{\bsnm{Teh}, \binits{Y.W.}}:
\bctitle{Set transformer: A framework for attention-based permutation-invariant
  neural networks}.
In: \bbtitle{Proceedings of the 36th International Conference on Machine
  Learning (ICML)}
(\byear{2019})
\end{bchapter}
\endbibitem

\bibitem[\protect\citeauthoryear{De~Raedt et~al.}{2007}]{de2007problog}
\begin{bchapter}
\bauthor{\bsnm{De~Raedt}, \binits{L.}},
\bauthor{\bsnm{Kimmig}, \binits{A.}},
\bauthor{\bsnm{Toivonen}, \binits{H.}}:
\bctitle{Problog: A probabilistic prolog and its application in link
  discovery.}
In: \bbtitle{IJCAI},
vol. \bseriesno{7},
pp. \bfpage{2462}--\blpage{2467}
(\byear{2007}).
\bcomment{Hyderabad}
\end{bchapter}
\endbibitem

\bibitem[\protect\citeauthoryear{Lapuschkin
  et~al.}{2019}]{lapuschkin2019unmasking}
\begin{botherref}
\oauthor{\bsnm{Lapuschkin}, \binits{S.}},
\oauthor{\bsnm{W{\"a}ldchen}, \binits{S.}},
\oauthor{\bsnm{Binder}, \binits{A.}},
\oauthor{\bsnm{Montavon}, \binits{G.}},
\oauthor{\bsnm{Samek}, \binits{W.}},
\oauthor{\bsnm{M{\"u}ller}, \binits{K.-R.}}:
Unmasking clever hans predictors and assessing what machines really learn.
Nature communications
\textbf{10}
(2019)
\end{botherref}
\endbibitem

\bibitem[\protect\citeauthoryear{Kwisthout}{2011}]{kwisthout2011mpe}
\begin{barticle}
\bauthor{\bsnm{Kwisthout}, \binits{J.}}:
\batitle{Most probable explanations in bayesian networks: Complexity and
  tractability}.
\bjtitle{International Journal of Approximate Reasoning}
\bvolume{52}(\bissue{9}),
\bfpage{1452}--\blpage{1469}
(\byear{2011})
\end{barticle}
\endbibitem

\bibitem[\protect\citeauthoryear{Petersen
  et~al.}{2021}]{petersen2021algorithmic_supervision}
\begin{bchapter}
\bauthor{\bsnm{Petersen}, \binits{F.}},
\bauthor{\bsnm{Borgelt}, \binits{C.}},
\bauthor{\bsnm{Kuehne}, \binits{H.}},
\bauthor{\bsnm{Deussen}, \binits{O.}}:
\bctitle{Learning with algorithmic supervision via continuous relaxations}.
In: \bbtitle{Advances in Neural Information Processing Systems (NeurIPS)}
(\byear{2021})
\end{bchapter}
\endbibitem

\end{thebibliography}
\newpage
\appendix

\section{Queries for Avoiding Infinite Loop } \label{loop}

We use four queries to test the performance of NEMESYS and ProbLog~\citep{de2007problog}.The four queries include one query which calls the recursive rule. The queries are:
\begin{align*}
\mathtt{query(path(a,a,[]))}.
\quad \mathtt{query(path(b,b,[]))}.\\
\mathtt{query(path(c,c,[]))}.\quad
\mathtt{query(path(a,c,A))}.
\end{align*}

\section{Differentiable Planning} \label{plan}

\begin{figure}
    \centering
    \includegraphics[width=0.243\columnwidth]{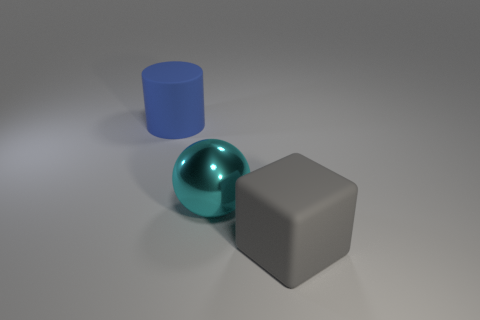}
    \includegraphics[width=0.243\columnwidth]{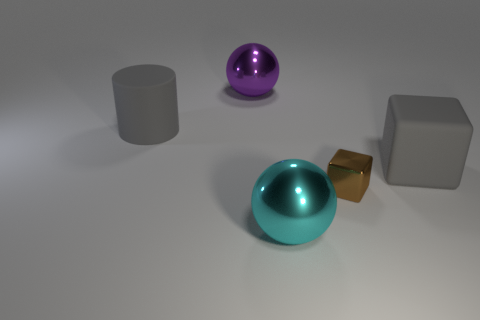} 
    \includegraphics[width=0.243\textwidth]{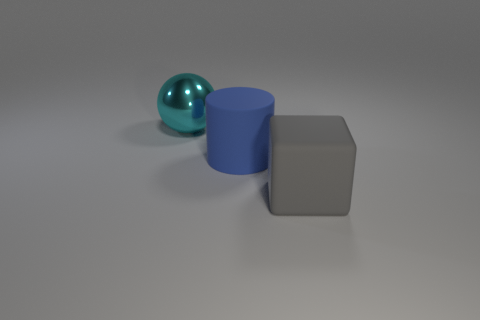}
    \includegraphics[width=0.243\textwidth]{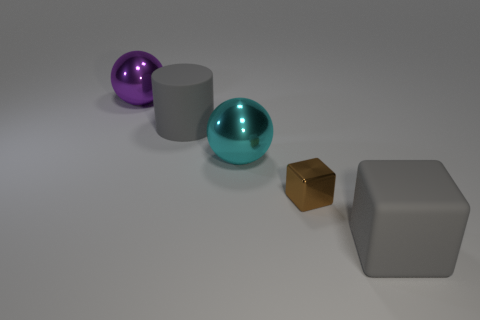}
    \caption{\textbf{Visual Concept Repairing: NEMESYS achieves planning by performing differentiable meta-level reasoning}. The left two images show the \emph{start} state, and the right two images show the \emph{goal} state. Taking these states as inputs, NEMESYS performs differentiable forward reasoning using meta-level clauses that simulate the planning steps and generate actions from \emph{start} state to reach the \emph{goal} state. (Best viewed in color)}
    \label{fig:additional_plan}
\end{figure}

We provide more planning tasks in Figure.~\ref{fig:additional_plan} with varying numbers of objects and attributes. Given the initial states and goal states, NEMESYS is asked to provide the intermediate steps to move different objects from start states to the end states. 

\section{Differentiable Parameter Learning Value Curve} \label{diff_do_value_curve}
\begin{figure}
    \centering
    \includegraphics[width=0.485\columnwidth]{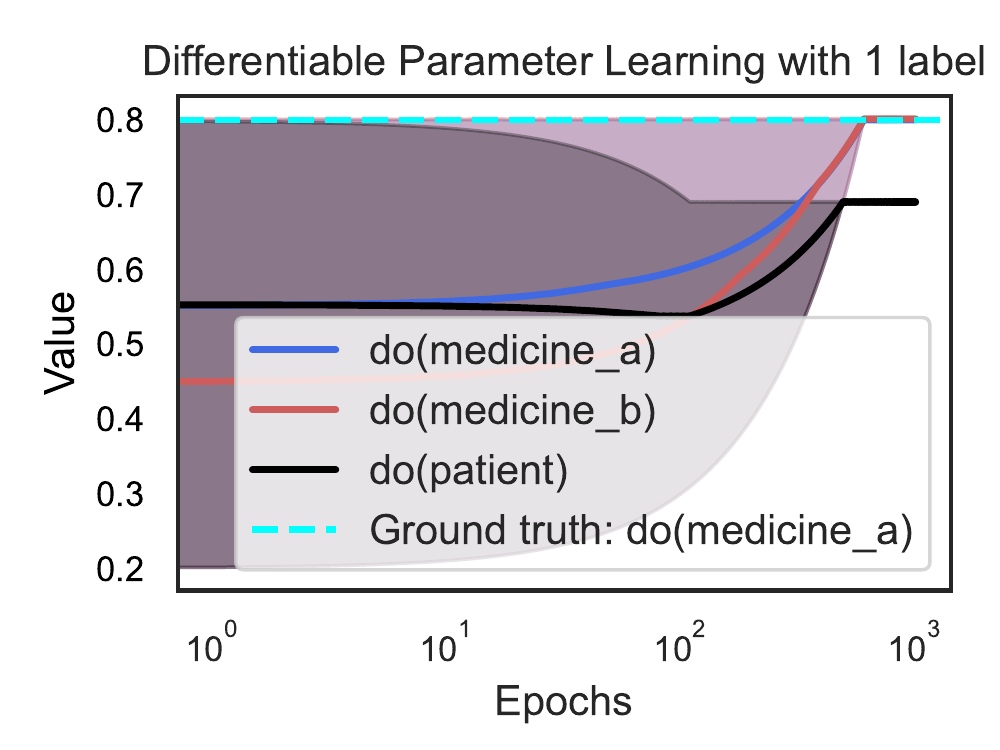}
    \includegraphics[width=0.485\columnwidth]{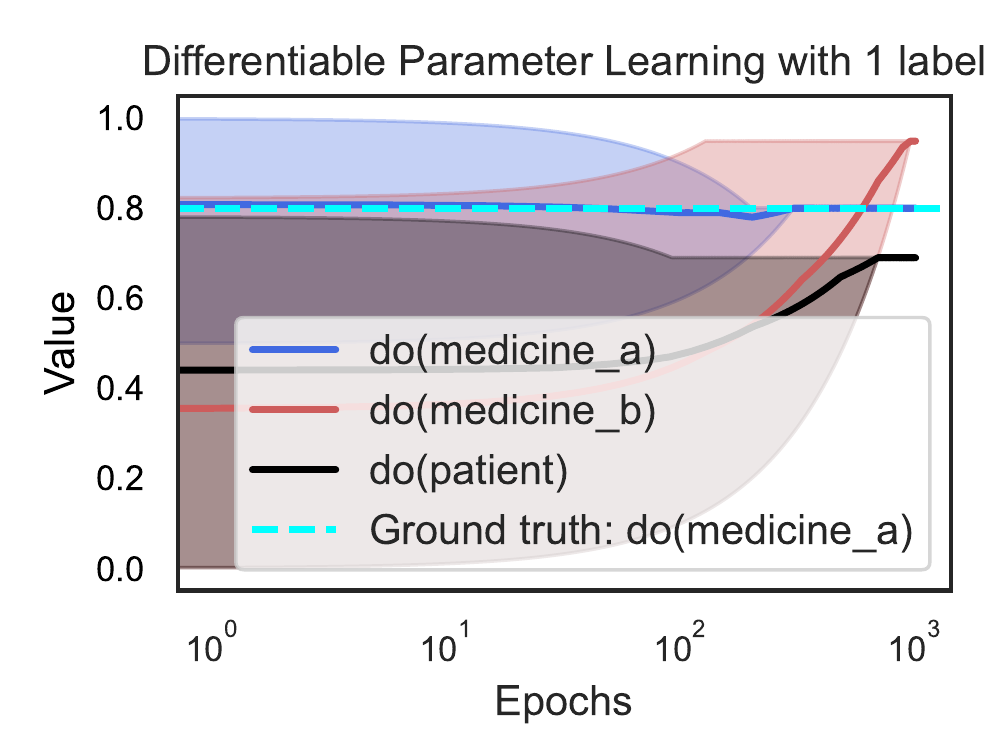} 

    \caption{\textbf{Value curve of three $\mathtt{do}$ operators during learning}. With three targets (right) and with one target (left). The curves are averaged on 5 runs, with shaded area indicating the maximum and minimum value. (Best viewed in color)}
    \label{do_value_curve}
\end{figure}

We also provide the corresponding value curve of these different $\mathtt{do}$ operators during learning in Fig.~\ref{do_value_curve}. In the experiment, we choose the $\mathtt{do}$ operator which achieves the lowest value as the correct value, thus in the experiment with three targets, we choose $\mathtt{do(medicine\_a)}$ with value $0.8$, which is exactly the ground-truth $\mathtt{do}$ operator with the correct number. 

\section{Multi-Task Adaptation} \label{multitask}

\begin{figure}[h]
    \centering
    \includegraphics[ width=\textwidth]{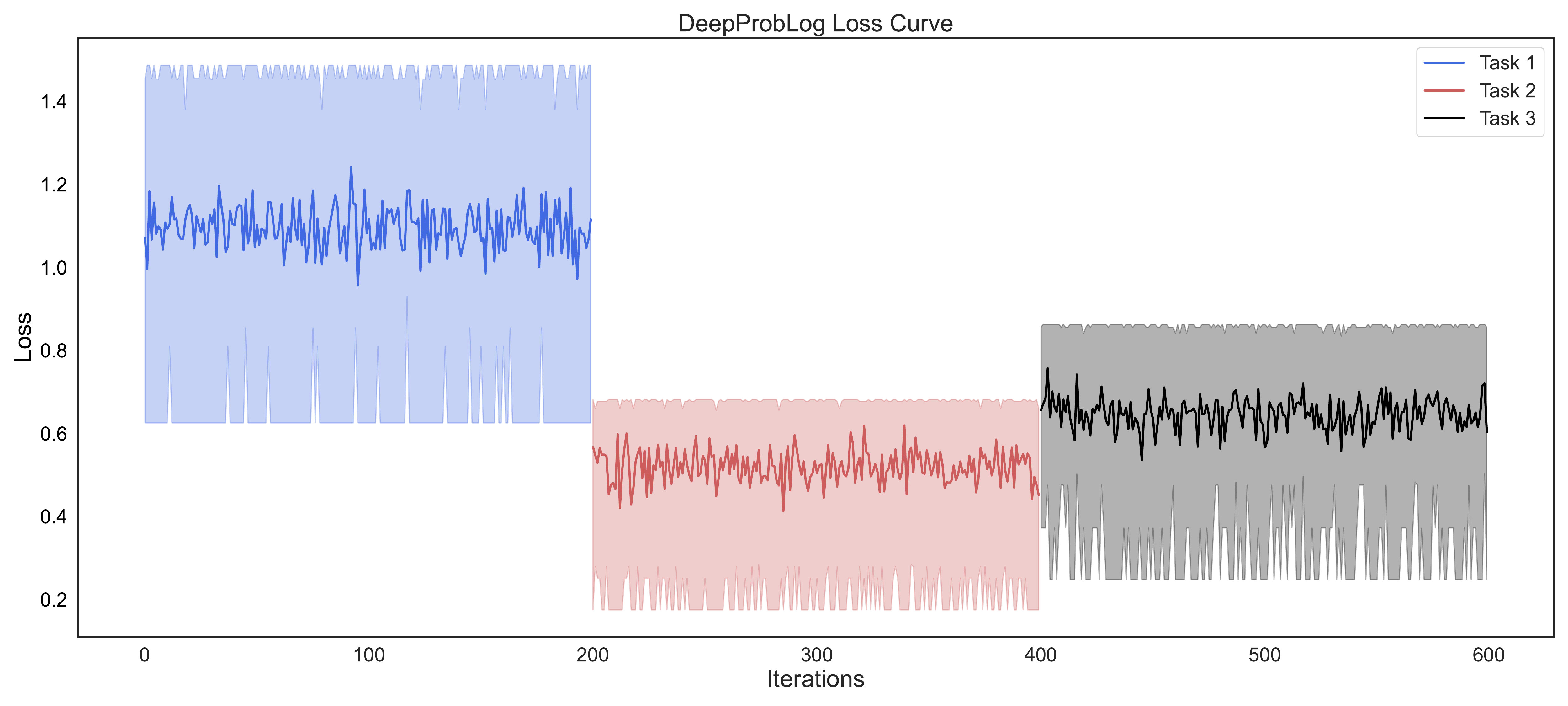}
    \caption{DeepProbLog~\citep{manhaeve2018deepproblog} is initialized using the same candidate meta rules (also with randomized meta rule weights as NEMESYS). The loss curve is averaged on five runs, with the shadow area indicating the minimum and maximum number of the five runs. (Best viewed in color)}
    \label{fig:multitask_problog}
\end{figure}

\begin{figure}
    \centering
    \includegraphics[trim=8 82 5 80, clip, width=\textwidth]{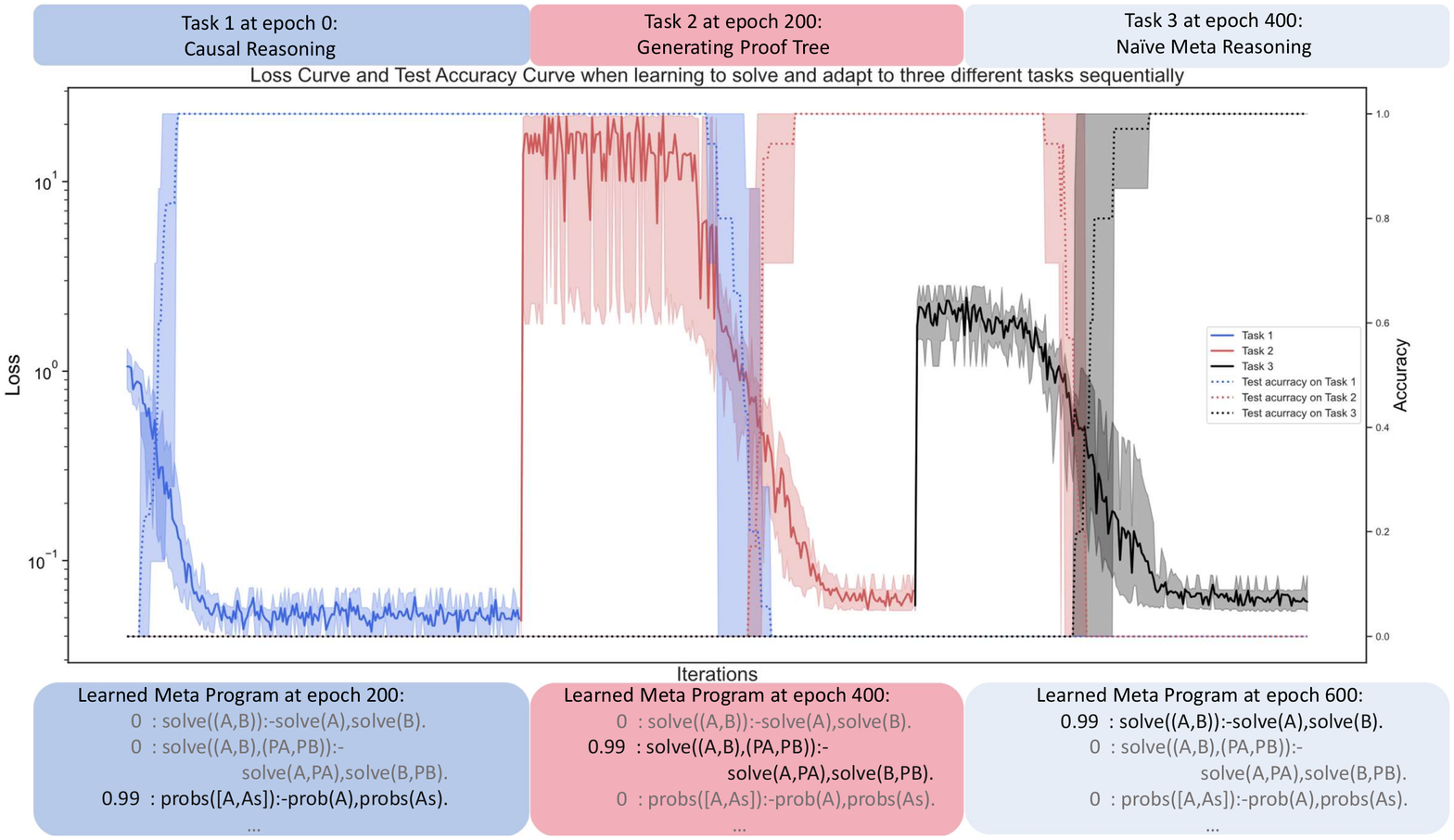}
    \caption{\textbf{Loss curve and accuracy curve of NEMESYS when learning to adapt to solve three tasks.} NEMESYS solves three different tasks (causal reasoning, generating proof trees and naive meta reasoning) sequentially (each task is represented by a unique color encoding). The loss curve (solid line) and accuracy curve (dashed line) are averaged on five runs, with the shadow area indicating the minimum and maximum number of the five runs. For readability, the learned complete meta program is shown in the text. (Best viewed in color)}
    \label{appendix:multi}
\end{figure}
We also compute the accuracy on the test splits of three tasks during the learning process (Fig.~\ref{appendix:multi} dashed line, color encoded). We choose DeepProbLog~\citep{manhaeve2018deepproblog} as our baseline comparison method in this experiment, however, learning the weights of (meta) rules is not supported in DeepProbLog framework, thus we randomly initialized the weights of the meta rules and compute the loss (Fig~\ref{fig:multitask_problog}). 

In this paragraph, we provide the meta program learned by NEMESYS in the experiment. The weights of meta rules are color coded to visually represent how their values evolve during the learning process (the weights are provided at iteration $\mathtt{200}$, $\mathtt{400}$ and $\mathtt{600}$), as illustrated in the accompanying Fig.~\ref{appendix:multi}. 
\begin{align*}
\\
&\hspace{1.5ex} \color{blue}0\ \hspace{1ex}\ \color{indianred}0.99\ \hspace{1ex}\ \color{black}0\ :
\mathtt{solve(A,B)}\texttt{:-}\mathtt{solve(A),solve(B).}\\
&\hspace{1.5ex}\color{blue}0\ \hspace{1ex}\ \color{indianred}{0.99}\ \hspace{1ex}\ \color{black}0\ :
\mathtt{solve(A)} \texttt{:-}\mathtt{clause(A,B), solve(B).}\\
&\hspace{1.5ex}\color{blue}0\ \hspace{1ex}\  \hspace{0.6ex}\ \color{indianred}0 \hspace{1ex}\ \color{black}0.99\ :\mathtt{solve((A,B),}\mathtt{(proofA,proofB))} \texttt{:-}\mathtt{solve(A,proofA),solve(B,proofB).}\\
&\hspace{1.5ex}\color{blue}0\ \hspace{1ex}\ \hspace{0.6ex}\ \color{indianred}0\ \hspace{0.25ex}\ \color{black}0.99\ :\mathtt{solve(A,}\mathtt{(A\texttt{:-}proofB))} \texttt{:-}\mathtt{clause(A,B), solve(B,proofB).} \\
&\color{blue}{0.99}\ \hspace{1ex}\ \color{indianred}0\ \hspace{1.5ex}\ \hspace{1ex} \color{black}0\ :
\mathtt{prob(Head)}\texttt{:-}\mathtt{assert\_probs((Head\texttt{:-}Body))},\mathtt{probs(Body).} \\
&\color{blue}{0.99}\ \hspace{1ex}\ \color{indianred}0\ \hspace{1.5ex}\ \hspace{1ex} \color{black}0\ :
\mathtt{probs([Body\_Atom,Body])}\texttt{:-}\mathtt{prob(Body\_Atom)},\mathtt{probs(Body)}. \\
&\color{blue}{0.99}\ \hspace{1ex}\  \color{indianred}0\ \hspace{1.5ex}\ \hspace{1ex} \color{black}0\ :
\mathtt{prob(Body\_Atom)}\texttt{:-}\mathtt{assert\_probs(Body\_Atom)}.  
\end{align*}

\end{document}